\documentclass[journal]{IEEEtran}
\usepackage[strings]{underscore}

\usepackage{url}

\usepackage{amsfonts}

\usepackage{amsmath}
\DeclareMathOperator*{\argmin}{arg\,min}

\usepackage{amssymb}

\usepackage[usenames, dvipsnames]{color}
\definecolor{orange}{RGB}{255,165,0}
\definecolor{brown}{RGB}{165,42,42}
\definecolor{purple}{RGB}{128,0,128}

\newcommand{\specialcell}[2][c]{%
  \begin{tabular}[#1]{@{}c@{}}#2\end{tabular}}
  
\usepackage[final]{changes} 
\setremarkmarkup{ \{\$#2\$\}}
\setauthormarkup{}
\definechangesauthor[name={Reviewer1}, color=orange]{r1}
\definechangesauthor[name={Reviewer2}, color=red]{r2}
\definechangesauthor[name={Reviewer3}, color=green]{r3}
\definechangesauthor[name={Reviewer4}, color=blue]{r4}
\definechangesauthor[name={mySelf}, color=cyan]{self}

\usepackage{empheq}

\ifCLASSINFOpdf
  \usepackage[pdftex]{graphicx}
  \graphicspath{{figures/}}
  \DeclareGraphicsExtensions{.jpeg,.png}
\else
\fi
\begin{document}
%
\title{Early Prediction for Physical Human Robot Collaboration in the Operating Room}
%
%
%

\author{Tian~Zhou,~\IEEEmembership{Student Member,~IEEE,}
        and~Juan~Wachs,~\IEEEmembership{Member,~IEEE}
\thanks{T. Zhou and J. Wachs are with the Department
of Industrial Engineering, Purdue University, West Lafayette, IN, 47906. e-mail: \texttt{\{zhou338, jpwachs\}@purdue.edu}.}
\thanks{Research supported by the NPRP award (NPRP 6-449-2-181) from the Qatar National Research Fund (a member of The Qatar Foundation). The statements made herein are solely the responsibility of the authors.}}

%
%

\markboth{To appear in \textit{Autonomous Robots}, special issue in Learning for Human-Robot Collaboration}%
{Shell \MakeLowercase{\textit{et al.}}: Bare Demo of IEEEtran.cls for IEEE Journals}
%



\maketitle

\begin{abstract}
To enable a natural and fluent human robot collaboration flow, it is critical for a robot to comprehend their human peers' on-going actions, predict their behaviors in the near future, and plan its actions correspondingly. Specifically, the capability of making early predictions is important, so that the robot can foresee the precise timing of a turn-taking event and start motion planning and execution early enough to smooth the turn-taking transition. Such proactive behavior would reduce human's waiting time, increase efficiency and enhance naturalness in collaborative task. To that end, this paper presents the design and implementation of an early turn-taking prediction algorithm, catered for physical human robot collaboration scenarios. Specifically, a Robotic Scrub Nurse (RSN) system which can comprehend surgeon's multimodal communication cues and perform turn-taking prediction is presented. The developed algorithm was tested on a collected data set of simulated surgical procedures in a surgeon-nurse tandem. The proposed turn-taking prediction algorithm is found to be significantly superior to its algorithmic counterparts, and is more accurate than human baseline when little partial input is given (less than 30\% of full action). After observing more information, the algorithm can achieve comparable performances as humans with a \textit{F1 score} of 0.90.
\end{abstract}


%
\IEEEpeerreviewmaketitle

\section{Introduction} \label{sec:intro}
%
%
%
%
\IEEEPARstart{T}{urn-taking} prediction is about the capability to comprehend the on-going task progress and  predict where, when and how to seize the next turn during multi-agent collaborations \cite{sacks_simplest_1974}. Fluent and natural turn-taking regulations would greatly increase team performances and lead to better social connections among team members \cite{sebanz_joint_2006,marsh_social_2009}. Robots, designed to work in close proximity with humans, need to have the capability to understand turn-taking events and plan their actions accordingly. A critical requirement for fluent turn-taking regulation is the capability to make decisions and preparations for turn-taking events beforehand \cite{sacks_simplest_1974}. Specifically, robots which perform physical interactions with humans need significant amount of time to plan and execute their motions \cite{canny_complexity_1988}, thus being able to make a prediction as early as possible would reduce human partners' waiting time. During collaboration, such early-prediction behavior would minimize mutual silence (both parties relinquish the turn) and mutual conflict (both parties attempt to seize the turn simultaneously), leading to a more synchronized turn-taking regulation \cite{cutler_analysis_1986}. \par

\begin{figure}[!t]
\centering
\includegraphics[width=3.2in]{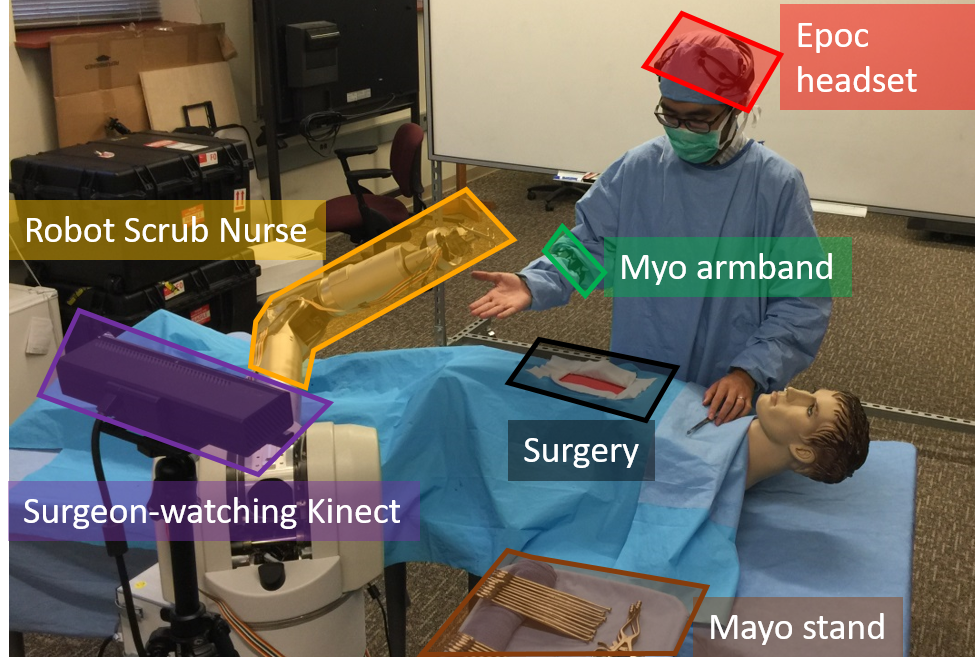}
\caption{System setup for the robotic scrub nurse. The surgeon is conducting a surgery (black) while the robotic nurse (\textcolor{orange}{orange}) picks up the requested instrument from mayo stand (\textcolor{brown}{brown}) and delivers to surgeon. The surgeon is monitored by Myo armband (\textcolor{green}{green}), Epoc headset (\textcolor{red}{red}) and Kinect (\textcolor{purple}{purple}) for turn-taking prediction.}
\label{fig:intro-system}
\end{figure}

The requirement for early turn-taking prediction stands out more clearly in high-risk and high-paced tasks like surgery. In the Operating Room (OR), the scrub nurse and the surgeon perform fast, accurate and highly coordinated turn-taking actions when exchanging surgical instruments. A nurse delivers surgical instruments to a surgeon based on explicit requests (e.g., uttering the words ``scissors'') and implicit requests expressed by body language (e.g., leaning forward, evoking a hand gesture or looking at an instrument). All these forms of expressions are used to inform the nurse ahead of time that it is his/her turn to continue. \added[id=r3,remark={r3c1:nurses also understand workflow and predict next instrument}]{Additionally, the nurses maintain a knowledge-base of common surgery procedures. Based on observing the current work-flow, the nurses can predict the most-likely next surgical operation and prepare instruments correspondingly. All such context knowledge about the surgical task helps nurses to better predict the timing (i.e., when) and the objects (i.e., what) of the next turn.} Such complex and coordinated turn-taking behaviors between surgeons and nurses are learned, acquired and executed precisely in the OR through experience and teams' practice. As robots are being introduced gradually to the OR to deal with nurses' shortage problem \cite{buerhaus2009recent, needleman2011nurse,zhou_early_2016}, they are expected to have the capability of performing fluent turn-taking actions as human nurses. \par

Such Robotic Scrub Nurse (RSN) will need to understand subtle verbal and non-verbal cues and infer the surgeon's intentions timely and accurately, in order to perform turn-taking actions correspondly. An illustration of the proposed RSN system is shown in Figure \ref{fig:intro-system}. To design a fully functional RSN, it is critical to develop a computational framework for early turn-taking prediction and that is the focus of this paper. 


\begin{table*}[!t]
\centering
\renewcommand{\arraystretch}{1.2}
\caption{Comparison of the proposed algorithm with state-of-the-art}
\begin{tabular}{|l|c|c|c|c|c|c|c|c|c|}
\hline
& \multicolumn{3}{c|}{Turn-taking timing range} & \multicolumn{4}{c|}{Modality} & \multicolumn{2}{c|}{Collaboration agents} \\
\hline
Paper & \specialcell{Early\\(0-40\%)} & \specialcell{Right-before\\(40\%-99\%)}  &	\specialcell{Just-in-time\\(99\%-100\%)}   & Gesture	& Gaze	&Speech	 & \specialcell{EEG/ \\EMG}	& \specialcell{Human-\\Human} &	\specialcell{Human-\\Robot} \\
\hline
Guntakandla \& Nielsen \cite{guntakandla_modelling_2015}  & & \checkmark & & & & \checkmark & & \checkmark & \\
\hline
De Kok \& Heylen \cite{de_kok_multimodal_2009} & & &\checkmark &\checkmark &\checkmark &\checkmark & &\checkmark & \\
\hline
Schlangen \cite{schlangen_reaction_2006} & &\checkmark & & & &\checkmark & & \checkmark& \\
\hline
Raux \& Eskenazi \cite{raux_optimizing_2012} & & &\checkmark & & &\checkmark & & \checkmark&\checkmark \\
\hline
Hart et al. \cite{hart_gesture_????} & & \checkmark& & \checkmark& \checkmark& & & \checkmark&\checkmark \\
\hline
Calisgan et al. \cite{calisgan_identifying_2012} & & &\checkmark &\checkmark &\checkmark & & & \checkmark&\checkmark \\
\hline
Chao \& Thomaz \cite{chao_timed_2012} & &\checkmark & &\checkmark &\checkmark &\checkmark & & \checkmark&\checkmark \\
\hline
Dumas et al. \cite{dumas_benchmarking_2009} & & &\checkmark &\checkmark & & &\checkmark &\checkmark & \\
\hline
Enrlich et al. \cite{ehrlich_when_2014} & & &\checkmark & & & & \checkmark& \checkmark& \checkmark\\
\hline
Heger et al. \cite{heger_eeg_2011} & & & \checkmark& & &\checkmark &\checkmark & &\checkmark	 \\
\hline
Bagci et al. \cite{kose-bagci_emergent_2008} & & & \checkmark& & &\checkmark & & &\checkmark \\
\hline
Matsusaka et al. \cite{matsuyama_towards_2015}& & &\checkmark & &\checkmark &\checkmark & & \checkmark&\checkmark \\
\hline
Mutlu et al. \cite{mutlu_footing_2009} & & & \checkmark& & \checkmark& & & &\checkmark\\
\hline
Yamazaki et al. \cite{yamazaki_precision_2008} & & & \checkmark& \checkmark& \checkmark& \checkmark& & \checkmark&\checkmark \\
\hline
\textbf{Ours} & \checkmark& & & \checkmark& \checkmark& \checkmark& \checkmark& \checkmark&\checkmark \\
\hline
\end{tabular}
\label{tab:literature} 
\end{table*}

\section{Related Work} \label{sec:related}
This section presents the related work about turn-taking analysis, in the perspective of 1) human-human turn-taking; 2) human-robot turn-taking; 3) automatic turn-taking recognition and 4) predictive turn-taking. To conclude this section, the innovation points are presented. \par 

\subsubsection{Human-human turn-taking} the analysis of turn-taking in human-human interactions has drawn attention from researchers with psychology, linguistics and engineering background. The research has focused on conversational tasks, where linguistic turn-ending cues such as pause duration \cite{schlangen_reaction_2006}, pitch levels \cite{ward_dialog_2010-1} and intonation \cite{gravano_turn-taking_2011}  have been identified. Nonverbal behaviors such as gaze and posture shifts \cite{padilha_nonverbal_2003} have also been studied during conversational exchanges, however, these are more difficult to spot during interaction.

\subsubsection{Human-robot turn-taking} in human robot interaction area, the CHARM project \cite{hart_gesture_????} studied nonverbal cues as key contributors to timing coordination among human collaborators. Its goal was to develop a turn-taking aware robot assistant to work alongside human workers in a manufacturing environment. Calisgan et al. \cite{calisgan_identifying_2012} studied the different types and occurrences of implicit communication cues as turn-taking “regulators” in assembly tasks. Also, timing in multimodal turn-taking interaction (i.e., speech, gaze, gesture) was investigated between humans and robots through a collaborative Towers of Hanoi challenge by Chao \& Thomaz \cite{chao_timed_2012}. \added[id=r3,remark={r3c2:literature of robot human handover}]{The naturalness of human robot turn-taking has also been studied in robot-to-human handover tasks, from the perspective of distinct handover poses \cite{cakmak2011using}, object affordances \cite{chan2014determining} and unambiguous approach angles \cite{unhelkar2014comparative}}. All the above-mentioned work focuses on observing and modelling turn-taking, without automatically recognizing them. \par

\subsubsection{Automatic turn-taking recognition} machine learning techniques have been applied to recognize turn-taking events automatically, mainly for spoken dialog systems. The speaker's end-of-turn is detected by an AI agent using Support Vector Machines \cite{arsikere2015enhanced}. Decision tree and its variants have also been used to detect turn-taking \cite{schlangen_reaction_2006,saito_estimating_2015,raux_optimizing_2008}. Such approaches have been extended to multimodal end-of-turn detection in multi-party meetings using Conditional Random Field \cite{de_kok_multimodal_2009}. \par 

\subsubsection{Predictive turn-taking} turn-taking modelling has to enable early prediction so that turn-taking decisions are made much earlier before the transition event occurs \cite{sacks_simplest_1974}. Given an observation of human actions, its meaning needs to be interpreted before the action is completely finished. For example, if an algorithm can only recognize the action meaning after it is fully conducted ($100\%$), there is no prediction involved and the algorithm degenerates to classification. While on the other hand, predicting the action type given $0\%$ of data would mean the earliest possible prediction (before the action even takes place). Heeman \& Lunsford \cite{heeman_can_2015} evaluated human's performance when predicting who will speak next, achieving an accuracy of $61\%$. The work presented by Hart et al., \cite{gravano_turn-taking_2011} achieved similar results using utterance-related cues and machine learning techniques. Conversely to utterance, gestures have also been studied as indicators for turn-taking. Early gesture recognition has been studied with dynamic time warping \cite{mori_early_2006} and naive Bayes \cite{escalante_naive_2016} previously. \par 

A comprehensive comparison of the proposed turn-taking prediction algorithm with the relevant literature is presented in TABLE \ref{tab:literature}.  The main scope of this table is to compare how early the turn-taking is analyzed (measured in percentage of full event), the involved modalities and the collaboration agents, as guiding criteria for the design of innovative turn-taking algorithms. This paper fits within the described criteria, and its innovation points are summarized as:

\begin{itemize}
\item Proposed a statistics-based feature selection process for raw sensor input.
\item Leveraged a Long Short-Term Memory network [9] for early turn-taking prediction.
\item Applied the Dempster-Shafer Theory of evidence for sensor fusion to increase robustness.
\item Developed a system which enables multimodal sensing for human robot collaboration in the Operating Room.
\end{itemize}

\added[id=r1, remark={r1c1:contribution not clear}]{This paper contributes to both the theory and implementation aspects of turn-taking. To the best of our knowledge, it is the first time that the concept of ``early turn-taking prediction'' is introduced, with a corresponding framework and implementation. The comprehensive experiments further validate the performance of the proposed framework.}

\section{Turn-taking in Homogeneous Humans Team} \label{sec:hh}
This section describes the procedure used to collect observations during turn-taking activities in pure human's team during a surgical task. The recorded observations were then used to guide the design of the early turn-taking algorithm. In the following, the surgical task setup (section \ref{sec:hh-task}), multimodal signal collection process (section \ref{sec:hh-signal}) and human-state annotations (section \ref{sec:hh-annotation}) are discussed.

\subsection{Surgical Task Setup} \label{sec:hh-task}
\begin{figure}[!t]
\centering
\includegraphics[width=3.2in]{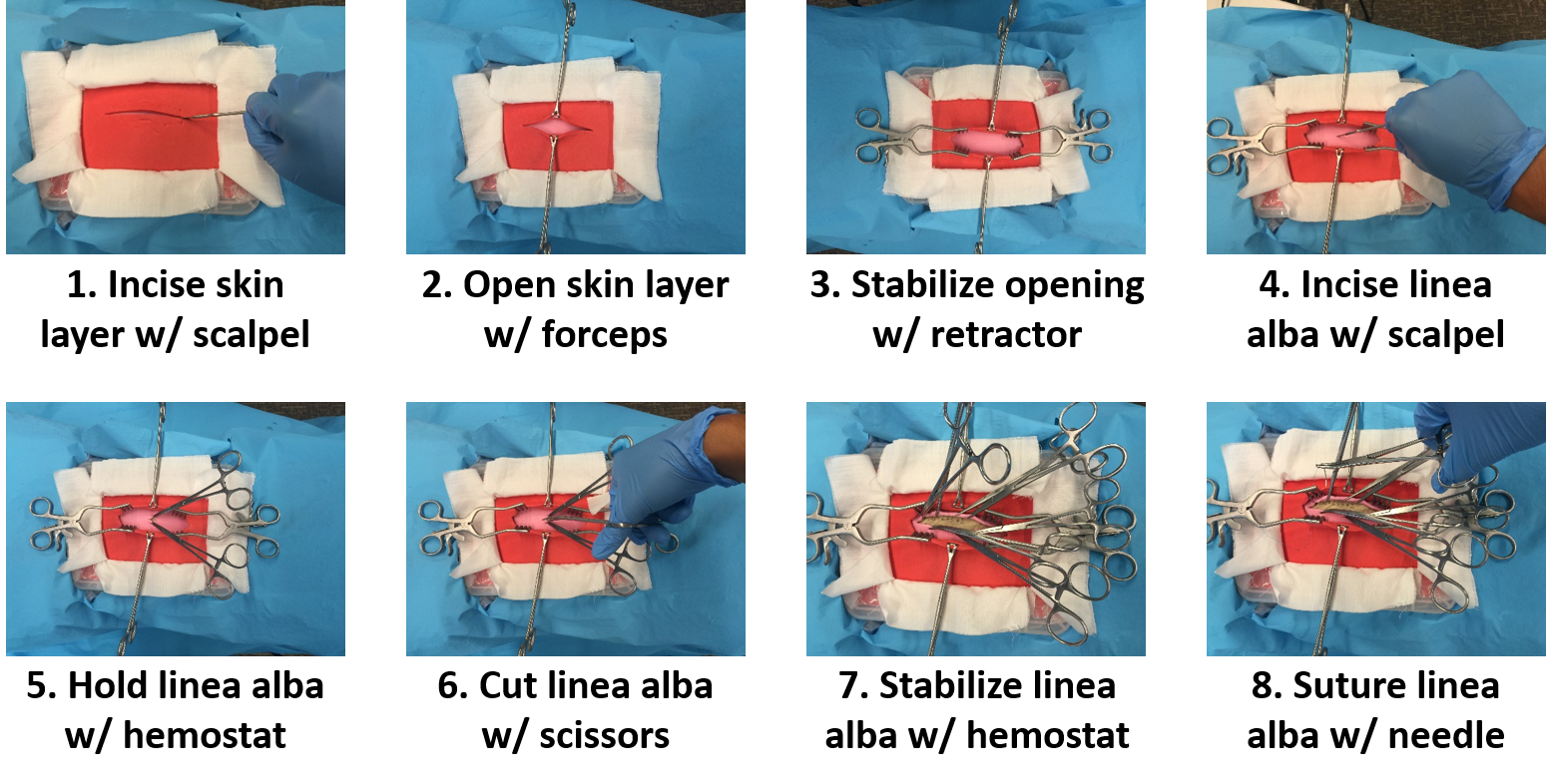}
\caption{Steps of the mock abdominal incision and closure surgical task with the surgical instruments used in each step. \added[id=r3, remark={r3c3: more explanation about surgical task}]{}}
\label{fig:hh-surgery}
\end{figure}

A simulation platform for surgical operations was used to capture turn-taking actions between surgeons and nurses, as shown in Figure \ref{fig:intro-system}. The platform includes a patient simulator and a set of instruments required to complete a mock surgical task of abdominal incision and closure \cite{martyak_abdominal_1976}. The detailed steps of the surgical task,\added[id=r3,remark={r3c3:more details of surgical task}]{ together with the surgical instruments needed for each step} are shown in Figure \ref{fig:hh-surgery}. In this task, the surgeon and the nurse collaborate by delivering and retrieving surgical instruments to complete the surgery successfully. \added[id=r4, remark={r4c7:simulation setup?}]{This simulation setup and the surgical task procedure have been used in our previous work (gesture robot nurse \cite{jacob_gestonurse:_2012} and tele-mentoring in the OR \cite{andersen2015virtual}).} \par

Participants were recruited to serve as surgeons. \added[id=r4, remark={r4c1:more details on instruction to participants}]{After signing the consent form, they were instructed about the steps of the mock abdominal incision and closure task through a video tutorial.} \added[id=r1, remark={r1c4:non-surgeons perform surgical task?}]{Then the participants performed a ``warm-up'' trial on the surgical simulator to increase familiarity with the task. Afterwards, each participant repeated the surgical task on the simulated setting five times in order to reach the expertise level required.} During the course of the surgical operations, the surgeons requested surgical instruments from the nurses, and handed back the used instruments. \added[id=r4, remark={r4c6:when does surgeon reach for instrument?}]{The surgeons would reach out for the instrument when it is presented by the nurse, and then continue to work on the task}. The surgeons were explicitly required to use verbal commands to request each instrument, \added[id=r4,remark={r4c4:surgeon use verbal commands?}]{in order to diminsh the effects of potential false nurse predictions}. \added[id=r4, remark={r4c3:noise in OR simulated?}]{The noise resulting from medical equipment and reverberations was not simulated.} In addition to the verbal commands, the surgeons' body, gaze and arm motions were all used together as implicit communication cues to trigger the nurses' actions, but no explicit request was given to the subjects about this. The nurse had to understand the surgeon's turn-taking communication cues (both implicit and explicit) in order to react according to the surgeon's expectations. \added[id=r4, remark={r4c2:subject might overreact to request instruments}]{Those implicit and explicit communication cues were collected through a set of sensors for the following turn-taking analysis, but the participants were not notified of any information of the specific sensing channels.} \added[id=r2, remark={r2c3:human predict robot motion? co-prediction?}]{The surgeon-nurse tandem forms a type of asymmetric collaboration, where the surgeon leads the task (i.e., a dominant agent) while the nurse mainly follows the task (i.e., a submissive agent). In this scenario, the focus is on enabling the \textit{follower} to predict the \textit{leader}'s turn-taking intention in order to collaborate efficiently. Thus, this paper only focuses on developing algorithms to enable robotic nurses to predict surgeons' turn-taking intentions, while ignoring the aspect of humans predicting robot motions, or co-prediction scenarios. } \par 

In this surgical task, the instrument request event was treated as the main turn-taking activity. The surgeon needs around 14 surgical instruments to finish one trial of the task, resulting in around 14 turn-taking instances). The instruments used are scalpel, hemostat, forceps, retractor, scissors and needle. Each participant repeated the surgical task 5 times. \added[id=r4, remark={r4c16:duration for full operation?}]{Each trial takes 3 to 7 minutes to finish, with subjective variations.} The study was approved by IRB (protocol number 1305013664), and the participants were recruited through emails and personal inquiries. In total $12$ participants were recruited with ages in range $20-31$ (mean = $25.7$, std = $2.93$). \added[id=r4,remark={r4c1:who are the participants?}]{The participants are all graduate students from the College of Engineering at Purdue University}, .

\subsection{Multimodal Signal Collection} \label{sec:hh-signal}
The communication cues expressed by the surgeon during the surgical task were recorded for further analysis. Three sensors were used to record those communication cues, namely Myo armband, Epoc headset and Kinect. An illustration of the captured multimodal signals is presented in Figure \ref{fig:hh-modality}. The details of each sensor channel is given below. 

\begin{figure}[!t]
\centering
\includegraphics[width=3.2in]{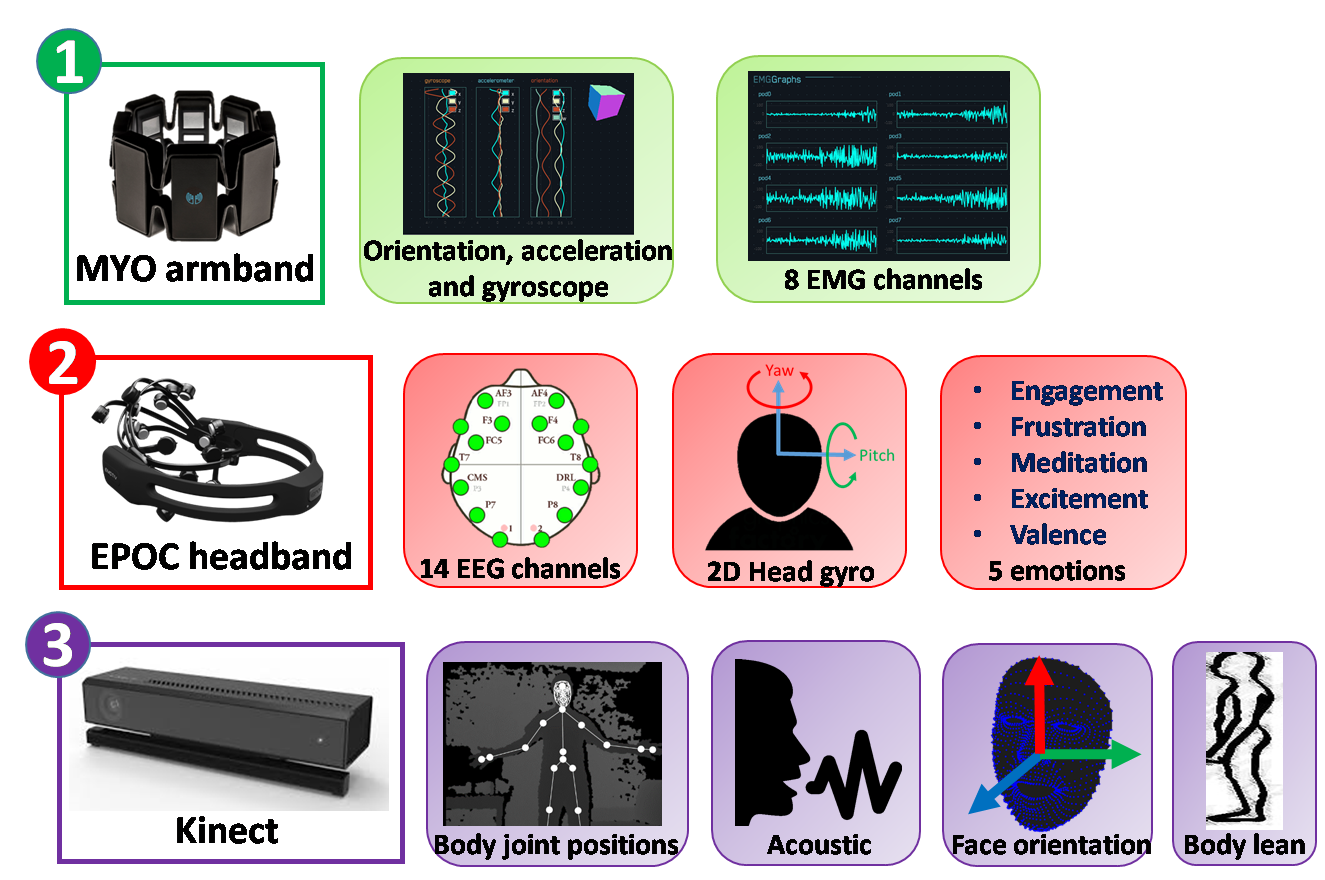}
\caption{Captured multimodal signals}
\label{fig:hh-modality}
\end{figure}

\subsubsection{Myo armband} a gesture capturing device worn on the forearm, capturing the motion and Electromyography (EMG) signals on the surgeon's dominant arm. The following information was recorded:
\begin{itemize}
\item Arm orientation (roll, pitch and yaw), 3D
\item Arm acceleration (xyz), 3D
\item Arm gyroscope (xyz), 3D
\item Arm muscle EMG signals, 8D
\end{itemize}

\subsubsection{Epoc headset} a brain-computer interface based on electroencephalography (EEG) technology. It is used to capture surgeon's head motions and EEG signals. The following data was recorded:
\begin{itemize}
\item Head EEG signals (AF3, F7, F3, FC5, T7, P7, O1, O2, P8, T8, FC6, F4, F8, AF42), 14D
\item Head gyro (pitch and yaw motion), 2D
\item Emotion classification (engagement, frustration, meditation, excitement and valence), 5D 
\end{itemize}

\subsubsection{Kinect} a motion sensing device. Joint, body and face tracking algorithms were used to extract participants' head poses, body postures and utterances. The following information was recorded:
\begin{itemize}
\item Face orientation (roll, pitch and yaw motion), 3D
\item Body postures (left-right leaning and forward-backward leaning), 2D
\item Left hand extension (vector from joint SpineMid to joint LeftHand), 3D
\item Right hand extension (vector from joint SpineMid to joint RightHand), 3D
\item Acoustic amplitude, 1D
\end{itemize}

\subsubsection{Synchronization} the real-time data from all three modalities was synchronized at a frame rate of $20 Hz$ and then concatenated together, forming a basic data level fusion. More advanced sensor fusion techniques will be introduced in section \ref{sec:dst}, for now the data level fusion only serves for data preparation and formatting purposes. After the data level fusion, for each time frame $t$, the fused sensor measurement (denoted as $\vec{r}_t$) consists of 50 values (addition of all the dimensions detailed above, denoted as $M$). Besides, the color images from Kinect were also recorded for annotation purposes (not included in $\vec{r}_t$). 

\subsection{Surgeon State Annotation} \label{sec:hh-annotation}
\begin{figure}[!t]
\centering
\includegraphics[width=3.2in]{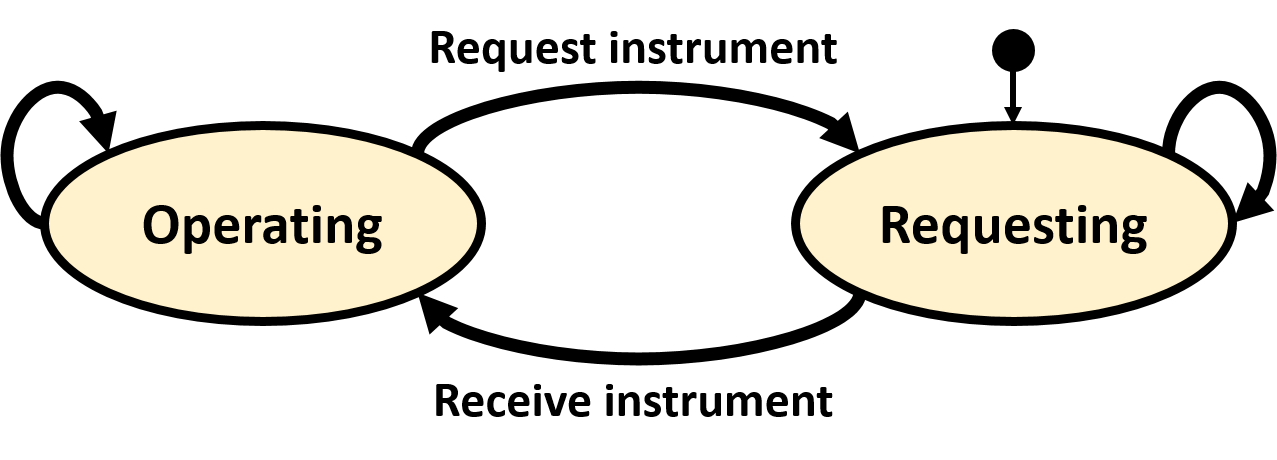}
\caption{Finite state machine to represent surgeon's states. The black dot indicates initial state. \added[id=self,remark={added two self-transition arrows to indicate that a state can remain itself}]{}}
\label{fig:hh-fsm}
\end{figure}

The recorded communication signal ($\vec{r}_t$) is obtained during periods when the surgeon is either focusing on the surgical operation or intending to request an instrument. These periods were further segmented into different surgeon states. \replaced[id=r1, remark={r1c5:FSM is very simple}]{It is assumed that the surgeon is always in one of the two states }{The two defined states of a surgeon are} (1) \textit{operating}: when the surgeon is engaged in current surgical operation \added[id=r4, remark={r4c10:definition of when turn-taking happens}]{and has no intention to relinquish the turn. This corresponds to the period when surgeon holds the current turn}; or (2) \textit{requesting}: the surgeon is requesting an instrument from the nurse. \added[id=r4, remark={r4c10:definition of when turn-taking happens}]{This corresponds to the period when surgeon is approaching the end of his/her turn, and wants to pass the turn to the nurse who will fetch and deliver the requested instrument. } A finite state machine to define the surgeon states is shown in Figure \ref{fig:hh-fsm}. In this scenario, instrument request events all happen during the switch from \textit{operating} state to \textit{requesting} state. \added[id=r2, remark={r2c2:human uncertainty like changing mind halfway}]{It is assumed by this paper that the human won't change his mind after switching to \textit{requesting} state, and won't return back to \textit{operating} state until receiving the requested surgical instrument.} Thus, the goal of the early turn-taking prediction algorithm is to predict the transition from \textit{operating} state to \textit{requesting} state as early as possible. \added[id=r1, remark={r1c5:FSM is very simple}]{Such binary end-of-turn detection approach (i.e., whether the human wants to keep or relinquish the turn) is a common practice in turn-taking analysis \cite{de_kok_multimodal_2009,guntakandla_modelling_2015,heeman_can_2015,arsikere_enhanced_2015}, and is also sufficient for the subsequent robot action decisions (i.e., engage interaction or not).}\par 

\begin{figure}[!t]
\centering
\includegraphics[width=3.2in]{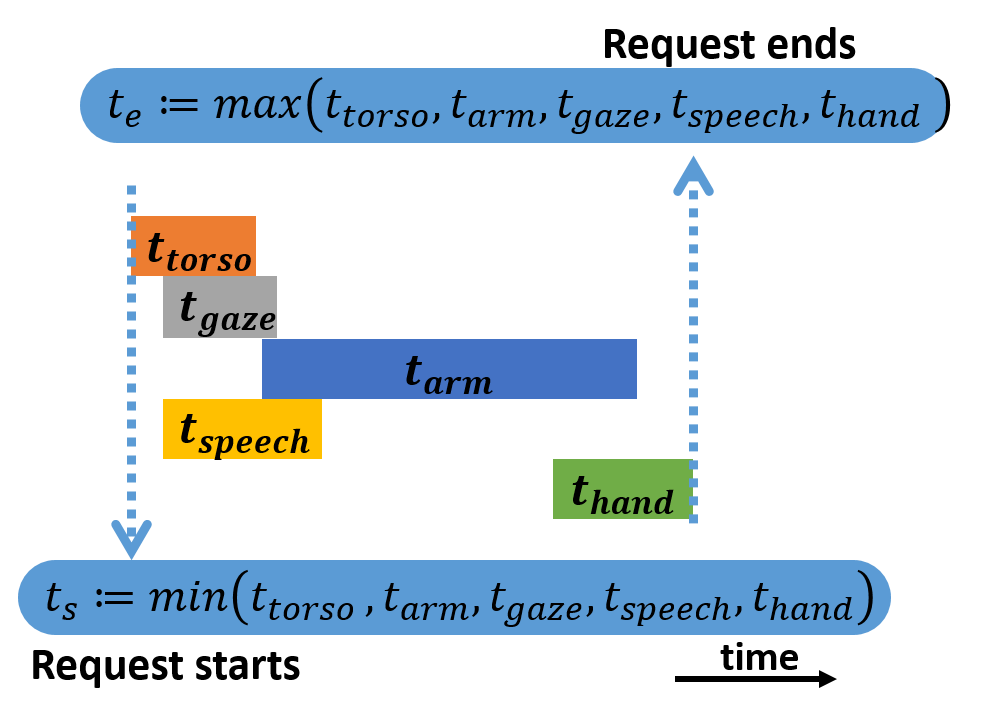}
\caption{Illustration of the annotation process for \textit{requesting} states. $t_*$ indicates the time period when modality $*$ is in active response}
\label{fig:hh-annotation}
\end{figure}

The recorded video was reviewed by humans to generate time indexes for each surgeon state. From the experiment recordings, it is observed that the \replaced[id=r4,remark={r4c8:do not make claims about surgeons in the OR}]{participants}{surgeons} use a combination of communication modalities to express their intent to request an instrument. Some modality starts early (e.g. changes in body stance) while others start late in the process(e.g. \replaced[id=r2, remark={r2c1:hand gesture as late modality}]{explicit hand gestures and specific verbal commands}{verbal requests}). Nurses can recognize the request intent as early as the earliest clue starts, or as late as the latest clue ends (for redundancy and cross-checking purposes). Therefore, the \textit{requesting} state is annotated to begin with the earliest clue ($t_s$), and end with the latest clue ($t_e$). In Figure \ref{fig:hh-annotation}, the determinatino of $t_s$ and $t_e$ is illustrated. More specifically, the modalities used to annotate videos are:
\begin{itemize}
\item Torso movement ($t_{torso}$): body stance was identified as one of the key communication cues in the OR \cite{moore_linguistic_2010}.
\item Gaze shift ($t_{gaze}$): gaze patterns were found to have high correlation with instrument handovers in OR \cite{mackenzie_hierarchical_2001}.
\item Arm movement ($t_{arm}$): preparatory arm movements were found to trigger the timing of turn-taking \cite{strabala_towards_2013}.
\item Speech command ($t_{speech}$): even though bringing many communication errors, verbal command in still one of the most common channels in the OR \cite{rabol_republished_2011}.
\item Hand gestures ($t_{hand}$): hand gestures are often used in the OR to request certain type of instrument \cite{gulasova_communication_????}.
\end{itemize}

For each video observation, the starting and ending time of the \textit{requesting} states were annotated, and the data between two consecutive \textit{requesting} states was considered to belong to the \textit{operating} state. A team member segmented all the videos and labeled the \textit{requesting} and \textit{operating} states, based on the above-mentioned criteria. Then, $10\%$ of randomly selected segments were labeled by a secondary team member. Inter-rater reliability showed \textit{almost perfect} agreement between the two sets of annotations with regard to the segment states (Cohen's $\kappa=0.95$) [41]. Overall, $846$ turn-taking instances (i.e., transitions from \textit{operating} state to \textit{requesting} state) were annotated and served as the basic data set for further experiments.

\section{Turn-taking Prediction} \label{sec:hr}
\begin{figure*}[!t]
\centering
\includegraphics[width=6in]{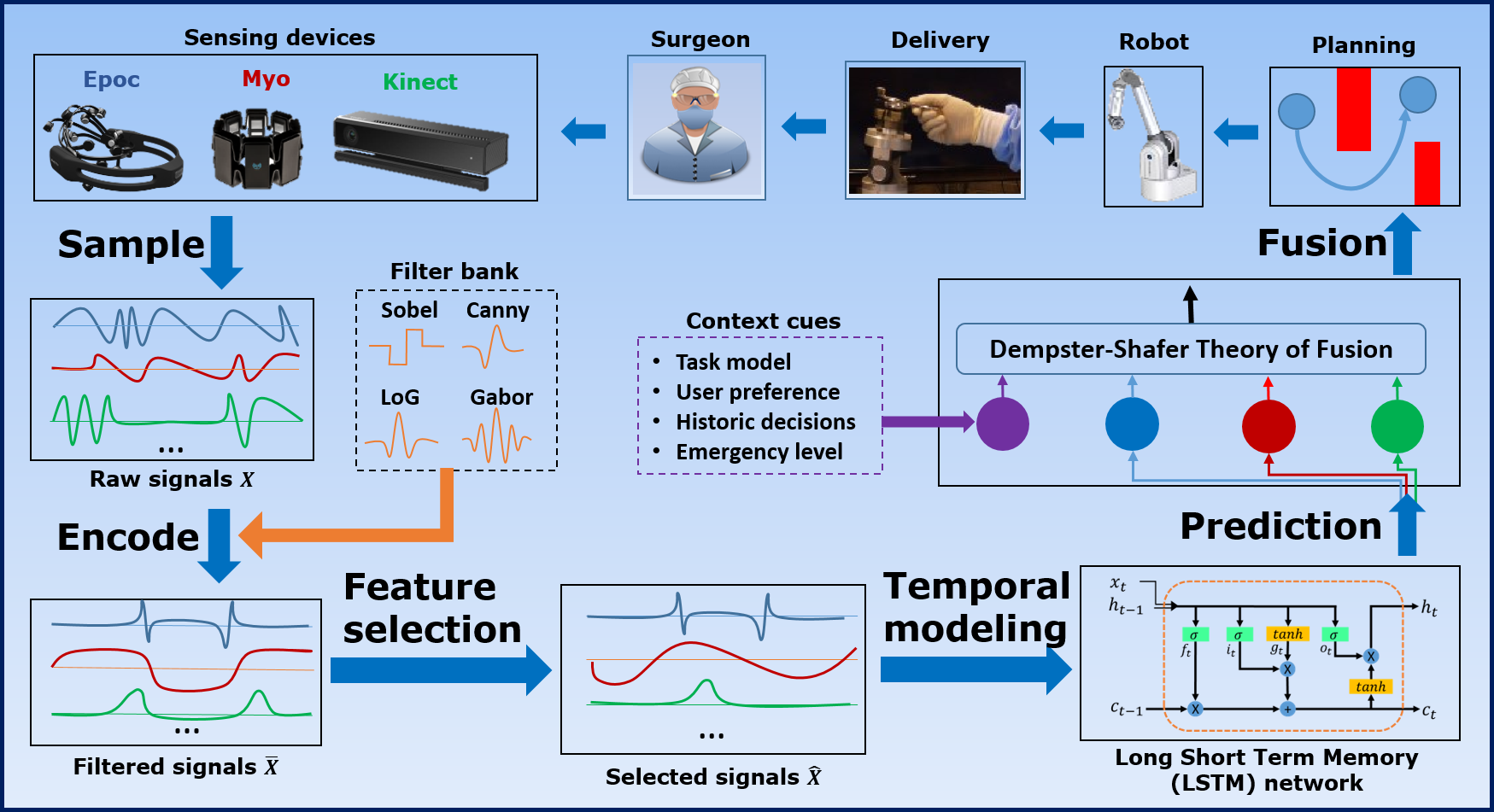}
\caption{Early turn-taking prediction framework \added[id=self, remark={expanded this figure to double column so that the text is more visible}]{}}
\label{fig:hr-architecture}
\end{figure*}

The main contribution of this paper is a computational framework for early turn-taking prediction, which can predict surgeon's turn-taking intentions before they are fully conceived. The proposed early turn-taking prediction framework is shown in Figure \ref{fig:hr-architecture}. The surgeon was monitored through three sensors, which are Epoc headset, Myo armband and Kinect sensor. The raw data was sampled and encoded, and the most relevant features were retained through a feature selection process. Then, the selected features were used for temporal modelling for turn-taking prediction. The prediction results from different resources were fused using the Demspter-Shafer method to achieve a final result. This result triggered the robot motion planning algorithm, which aims to pick up and deliver the surgical instrument to the surgeon at the right time. This paper mainly focuses on the turn-taking prediction and fusion part, while neglecting the robot motion planning aspect. However, the entire closed-loop process flow was presented here for integrity and illustration purposes. In the reminder of this section, detailed descriptions of channel preprocessing (section \ref{sec:hr-channel}), feature construction and selection (section \ref{sec:hr-feature}) and temporal modelling (section \ref{sec:hr-early}) are presented.

\subsection{Channel Preprocessing} \label{sec:hr-channel}
The first step in the signal processing pipeline was noise reduction. The Exponentially Weighted Moving Average (EWMA) method was used, which is a common noise reduction technique for time-series data \cite{lucas_exponentially_1990}. This filter was applied to smooth the raw signal $\vec{r}_t$:

\begin{equation}
\label{eq:ewma}
\vec{s}_t = \alpha \vec{r}_t + (1-\alpha)\vec{s}_{t-1}, t \in [1,L]; \vec{s}_0 = \vec{r}_0
\end{equation}
where $\vec{r}_t$ is the raw sensor measurement at time $t$,  $L$ is the discrete length of signal and $\vec{s}_t$ is the filtered measurement at time $t$. The smoothing was applied to each of the $M$ channels of $\vec{r}_t$. The weighting parameter $\alpha$ controls the relative importance of raw measurement, which was empirically set to $0.2$ for best performance in our environment. \par 

The smoothed signal $\vec{s}_t$ was then normalized to have zero mean and unit variance in each of the $M$ channels, following $\vec{x}_t\{c\}= (\vec{s}_t \{c\}-\mu_c)/\sigma^2_c$, where $\mu_c$ and $\sigma^2_c$ are the global mean and variance of channel $c$ in signal $\vec{s}_t$. Here the statistics $\mu_c$ and $\sigma^2_c$ were calculated based on $\vec{s}_t \{c\}$ from both states (\textit{requesting} and \textit{operating}), hence the name global. Such approach preserves any offset between data of different states, while at the same time enforces the multimodal signals to be in similar magnitudes for further comparison and aggregation purposes. \par 

For each time stamp $t$, the smoothed and normalized signal $\vec{x}_t$ consists of $M$ values. The collection of $\vec{x}_t$ within a given time window $i$ (e.g., $t\in[t_s^i,t_e^i]$) is cumulatively denoted as segment $\tilde{X}_i$. $\vec{x}_t$ in time window $i$ was stacked together to form a matrix representation $\tilde{X}_i \in \mathbb{R}^{L_i \times M}$, where $L_i$ is the length of segment $i$ (i.e., $L_i=t_e^i-t_s^i$) and $M$ is the dimension of raw data (i.e., $M=50$). For each stacked segment $\tilde{X}_i$, a label $y_i \in \{0,1\}$ was assigned to indicate whether segment $\tilde{X}_i$ belongs to \textit{requesting} state (\(y_i=1\)) or \textit{operating} state (\(y_i=0\)). The combination $\{\tilde{X}_i,y_i \}$ forms the initial data set for turn-taking analysis, which aims to infer $y_i$ from $\tilde{X}_i$.

\subsection{Feature Construction and Selection} \label{sec:hr-feature}
To construct a compact yet relevant feature representation, a feature construction and selection process was carried out. \added[id=r1,remark={r1c7:confusion about feature construction and selection}]{The basic principle is that the raw signal was first convolved with a set of filters to characterize different patterns in the original signal (i.e., feature construction stage). Then through correlation analysis, the most relevant features were selected for the subsequent analysis (i.e., feature selection stage). These two stages are described in more detail in the following. }\par 

\added[id=r1,remark={r1c7:confusion about feature construction and selection}]{\textbf{Feature construction:}} the signal after channel preprocessing (i.e., $\tilde{X}_i$) was temporally encoded by convolving with a set of filter banks. The set of filter banks $\{F_q | q=1, \ldots, Q\}$ includes $Q$ different filters to encode different temporal characteristics of the original signal. The filter bank includes the identity transformation, Sobel operator (window size \(3\)), Canny edge detector (approximated with derivative of Gaussian), Laplacian of Gaussian (\textit{LoG}) and Gabor filter (gabor1 with central frequency \(100 Hz\) and phase \(0\) and gabor2 with central frequency of \(100 Hz\) and phase $\pi/2$). For each segment $\tilde{X}_i \in \mathbb{R}^{L_i \times M}$, each of its $M$ channels was convolved with all $Q$ filters from the filter bank, resulting in $\bar{X}_i \in \mathbb{R}^{L_i \times MQ}$. \par 

\added[id=r1,remark={r1c7:confusion about feature construction and selection}]{\textbf{Feature selection:}} the features constructed from the $MQ$ channels (i.e. $\bar{X}_i$) showed high redundancy and hence only those channels highly correlated with the turn-taking labels were kept. Specifically, the $m$ most salient features were selected out of $MQ$ encoded features ($m \ll MQ$), based on a statistics-based approach. This method is inspired by Morency et. al, \cite{morency_context-based_2008}, with the major difference that the previous work requires manually annotated discrete inputs, while in this case the input data can have continuous values and is automatically encoded. This extension from discretely valued inputs to continuously valued input can increase the applicability of the feature selection algorithm, and also saves time and effort in manual annotation. \par 

The feature selection process works as the following. First, each of the $MQ$ continuous signals was converted into binary representations using clustering (\textit{K}-means with $2$ clusters and each cluster represents a binary level). Then, a $\chi^2$ test was carried out between each binarized signal and the ground truth label. The $MQ$ features were then ranked based on the test statistics of the corresponding $\chi^2$ test, and the $m$ features of the largest test statistics were retained as the optimal feature set, represented as $X_i \in \mathbb{R}^{L_i \times m}$. This representation includes the optimal feature set after being temporally encoded. 

\subsection{Early Prediction} \label{sec:hr-early}
We propose the usage of Long Short-Term Memory (LSTM) \cite{hochreiter_long_1997}, a type of recurrent neural network, for early turn-taking prediction. For comparison purposes, the Multi-Dimensional Dynamic Time Warping (MD-DTW) \cite{ten_holt_multi-dimensional_2007} is also tested. The DTW serves as a baseline and a representative of traditional temporal modelling algorithms.  

\subsubsection{LSTM} 
\begin{figure}[!t]
\centering
\includegraphics[width=3.2in]{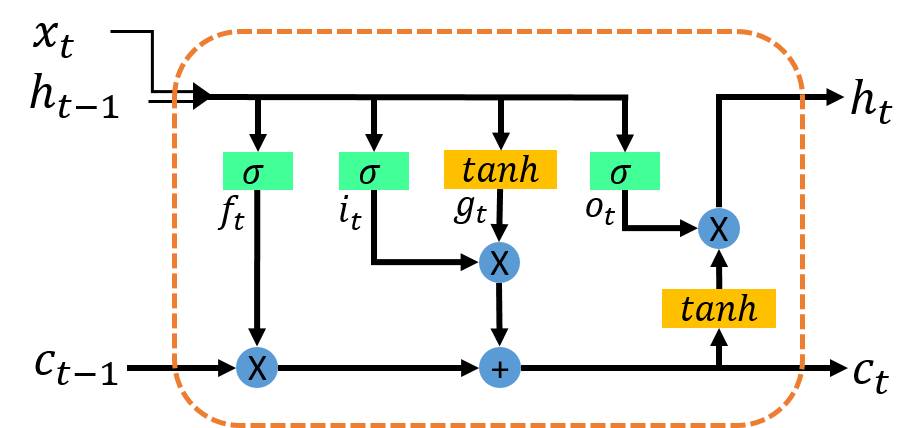}
\caption{Basic cell structure of LSTM \added[id=self,remark={a more colorful image for LSTM cell, looks nicer}]{}}
\label{fig:hr-lstm}
\end{figure}

it is a recurrent neural network architecture which has been successfully applied to handwriting recognition \cite{graves_novel_2009} and emotion recognition \cite{wollmer_lstm-modeling_2013}, among other applications. This network structure has the intrinsic capability to automatically extract spatial-temporal patterns. Even though there are many different recurrent neural network structures (e.g. Gated Recurrent Unit \cite{cho_learning_2014}), a recent comparison finds out that most popular variants perform similarly \cite{greff_lstm:_2015}. Thus, we decided to choose LSTM as the sequence modelling network for our scenario. The basic structure of a cell of LSTM is shown in Figure \ref{fig:hr-lstm}. \par 

The LSTM structure can be described by a set of formulas:
\begin{equation} \label{eq:lstm}
\begin{split}
f_t &= \sigma(W_{f}[x_t,h_{t-1}] + b_f) \\ 
i_t &= \sigma(W_{i}[x_t,h_{t-1}] + b_i) \\ 
g_t &= tanh(W_{g}[x_t,h_{t-1}] + b_g) \\
o_t &= \sigma(W_{o}[x_t,h_{t-1}] + b_o) \\ 
c_t &= f_t \odot c_{t-1} + i_t \odot g_t \\
h_t &= o_t \odot tanh(c_t) \\
y_t &= softmax(W_y h_t)
\end{split}
\end{equation}

where $x_t$ is the input temporal sequence at time $t$ (corresponding to a row in $X_i \in \mathbb{R}^{L_i \times m}$ as defined above), $h_{t-1}$ is the output of the memory cell at time $t-1$. $W_{f},W_{i},W_{g},W_{o}$ are weight matrices, and $b_f,b_i,b_g,b_o$ are bias terms respectively. $\sigma$ denotes a \textit{sigmoid} function, $tanh$ indicates the hyperbolic tangent function and $\odot$ denotes an element-wise multiplication. $[x_t,h_{t-1}]$ denotes the concatenation of vector $x_t$ and $h_{t-1}$. $f_t,i_t,g_t,o_t,c_t,h_t$ are forget gate, input gate, candidate gate, output gate, cell unit and hidden state respectively. The memory unit $c_t$ is generated by the coupling between the input and forget gates. LSTM can model long-term temporal dependencies because the memory unit can selectively ``remember'' or ``forget'' past information. The strategy to open or close gates is embedded in the learned weights and biases. \par 

During the training stage, the segment-label pairs $\{(X_i,y_i) | X_i \in \mathbb{R}^{L_i \times m}, y_i \in \{0,1\}, i \in [1, N_0+N_1] \}$ from both \textit{requesting} (in total $N_1$ instances) and \textit{operating} states (in total $N_0$ instances) were introduced into the network. The learning algorithm then updates the weights and biases for each gate, by minimizing the softmax cross entropy between the ground truth and the predicted labels. The Adam optimizer \cite{kingma_adam:_2014} was used to iteratively estimate the parameters of the network, based on stochastic gradient descent algorithm (SGD). \par 

During the testing stage, a given unknown sequence $X_k$ is presented to the network. The memory cell output of the last time step $h_t$ is multiplied by $W_y$ and then applied the \textit{softmax} function to determine the model output $\hat{y}_k\in \{0,1\}$, which is the predicted label for segment $X_k$.

\subsubsection{Multi-Dimensional Dynamic Time Warping (MD-DTW)} DTW is one of the most traditional and successful temporal modelling algorithms, and has been successfully applied to speech recognition \cite{abdulla_cross-words_2003}, early gesture recognition \cite{mori_early_2006} and robot trajectory navigation \cite{vakanski_trajectory_2012}. In \cite{ten_holt_multi-dimensional_2007}, 1D-DTW was extended to a multi-dimensional case (MD-DTW) and the superiority of MD-DTW over any 1D-DTW systems was shown. \par 

In this scenario, the MD-DTW algorithm \cite{ten_holt_multi-dimensional_2007} was applied with 1-norm as the distance measurement for two multi-dimensional signals. A nearest-neighbor classification scheme is used, which assigns the label of the closest example to the unknown sequence. \par 

The training stage consists of finding the most representative instance (known as the \textit{template}) for each surgeon state. Take the \textit{requesting} state as an example $(y_i=1)$. The selection of the \textit{template} for this state (denoted as $X_*^1$) is based on within-group consensus. The instance $X_i$  which has the least cumulative DTW distances with the remaining instances of the same state was chosen as the \textit{template}, i.e:

\begin{equation}
\label{eq:dtw_train}
X_*^1 = \argmin_{i=1}^{N_1} \sum_{j,j\neq i}^{N_1}{DTW(X_i, X_j)}
\end{equation}

\added[id=r4,remark={r4c15:why only two templates?}]{The same procedure is repeated for the \textit{operating} state and the resultant \textit{template} is denoted as $X_*^0$. This pair of  \textit{templates} $\{X_*^1,, X_*^0\}$ was selected based on the training instances from all the operation steps, and can generalize to novel operations/surgeries.}

During testing stage, for a given unknown sequence $X_k$, its DTW distance with the \textit{template} pairs $\{X_*^1, X_*^0\}$ is calculated. Then the label associated with the smaller distance is chosen as the prediction $\hat{y}_k$ for sequence $X_k$, i.e.:
\begin{equation}
\label{eq:dtw_test}
\hat{y}_k = \argmin_{y\in\{0,1\}} {DTW(X_k, X_*^y)}
\end{equation}

\section{Decision Fusion with Dempster-Shafer Theory} \label{sec:dst}
\added[id=r4,remark={r4c33:add a more explicit introduction}]{Turn-taking intent is expressed through several modality channels, together with additional context knowledge. This section discusses how to fuse the information from multiple channels together in order to achieve a robust and comprehensive turn-taking reasoning.}

Fusion can happen at data level, feature level and decision level. A decision-level fusion technique was utilized due to its low computational cost and robustness. First, independent classifiers were trained using information from different resources, and then the outputs of these classifiers were combined together using weighted average. Common approaches to find the weights include grid search \cite{hinton2010practical}, random search \cite{bergstra2012random} and supervised retraining on confidence outputs \cite{mittal_hand_2011}. However, the searching approaches are time-consuming and suffer from local minimum. Supervised retraining approach cannot generalize to scenarios where modalities are missing or added. Instead, the Dempster-Shafer Theory (DST) was resorted for decision fusion. \par 

DST provides a framework to combine degree of beliefs derived from independent evidence channels. The DST framework is easily extendable and allows for incremental modality addition or subtraction. Also, it provides a measurement of uncertainty levels and modality disagreement. \added[id=r1,remark={r1c9: why DST?}]{Compared to other uncertainty reasoning frameworks such as fuzzy logic \cite{klir1995fuzzy}, DST has the advantage of preserving uncertainty levels from each evidence channel and providing a boundary for the certainty levels of the final belief \cite{harmanec1994measuring}. Due to DST's wide applicability,} it has been used for decision fusion in visual tracking \cite{li_visual_2013}, human activity recognition \cite{mckeever_activity_2010} and robot localization  \cite{hughes1992ultrasonic}, to mention a few examples. \par 

In this scenario, the DST was used to accumulate confidences from both spatial domain (across different modalities and context knowledge) and temporal domain (across consecutive time frames), similar to the spatial-temporal weighted Dempster-Shafer scheme \cite{li_visual_2013}. The major difference is that our approach in addition considers contextual knowledge about task progress. 

\subsection{DST formulation} \label{sec:dst-formulation}
\added[id=r4,remark={r4c34:add a more explicit introduction}]{DST is considered as a generalization of Bayesian methods, with the difference that Bayesian method only assigns weights (i.e., probabilities) to each individual state of the system, while DST assigns weights (i.e., belief) to each combinatory propositions of the system states. Such a generalization grants DST more flexibility in modelling uncertainties and fusing beliefs from multiple modalities \cite{wu2002sensor}.}

Let $\epsilon=\{\epsilon_1,\epsilon_2,...,\epsilon_n \}$ represents the $n$ possible states of the system under consideration. The power set $2^\epsilon$ contains all subsets of $\epsilon$ and thus represents all propositions about the actual state of the system. The DST theory then assigns a \textit{Basic Belief Assignment} (\textit{BBA}) $b$, to each element of the power set $2^\epsilon$, i.e., $b:2^\epsilon \rightarrow [0,1]$ such that $b(\emptyset)=0$ and $\sum_{A\in 2^\epsilon} b(A) = 1$. For state $A \in 2^\epsilon$, its \textit{BBA} value $b(A)$ expresses the evidence to support the claim that the actual system state belongs to $A$. The DST theory also provides a framework to combine confidences from different sources, known as the \textit{Dempster's Rule of Combination} (DRC). It calculates a joint belief $b_{1,2}(\cdot)$  from two independent beliefs $b_1 (\cdot)$ and $b_2 (\cdot)$, according to:

\begin{equation}
\label{eq:drc}
b_{1,2}(A) = \frac{1}{1-K} \sum_{B \cap C = A} b_1(B)b_2(C)
\end{equation}

where $K= \sum_{B \cap C = \emptyset} b_1 (B) b_2 (C)$ is a measure of disagreement between the two beliefs $b_1 (\cdot)$ and $b_2 (\cdot)$. A large $K$ value implies strong disagreement between the two beliefs regarding the actual state of the system. 

\subsection{DST in Turn-taking Prediction} \label{sec:dst-usage}
We created a \textit{BBA} function from each modality available, and then used DRC to combine the evidence from all the \textit{BBA} dimensions to reach a single final decision. The construction of the \textit{BBA} for each modality is based on the output of each classifier. For LSTM, the output of memory cell of the last time step $h_t$ is multiplied by $W_y$ and then transformed by the \textit{softmax} function, i.e. \textit{softmax}($W_y h_t$). The result is a vector of values in the range $[0,1]$ that add up to $1$. This vector is used to initialize the \textit{BBA} values, by using the corresponding $i_{th}$ entry in the vector for $b(A=\{y_i\})$. \par 

In our system, three sensors were used to capture human communication cues, namely Epoc headset ($b_e$), Kinect ($b_k$) and Myo armband ($b_m$). A LSTM network is trained individually on features corresponding to each sensor. During testing, the output from each sensor is used to construct each \textit{BBA} function, and finally a joint belief function $\tilde{b}$ is calculated by combining individual \textit{BBA} functions using DRC. The ultimate decision is made solely based on the joint belief function $\tilde{b}$. \par 

The DRC fusion rule provides a robust and flexible framework for spatial-temporal confidence fusion. For example, in situations where one of the sensors is not available due to potential hardware failures, the DRC combination rules can still accumulate confidences from the working sensors and make a final decision. \par 

To make a final decision $\tilde{b}^t$ for time frame $t$, we used 1) the individual \textit{BBA} based on the output of the LSTM network for each modality, denoted as $b_e^t, b_k^t$ and $b_m^t$; 2) the final BBA from the previous time frame $\tilde{b}^{t-1}$ and 3) a context BBA $b_c^t$ which characterizes the current context cues. For now, the context cue $b_c^t$ only characterizes the task progress. \replaced[id=r4, remark={r4c18:confusion about context variable}]{The task progress is described by the amount of time past since the beginning of current operation, normalized by the total duration of this operation.} {The task progress is described by the amount of time past since the beginning of current operation (in percent of the total duration)}. We assume that the surgeon's intentions of requesting an instrument are expressed in a stronger manner as time goes on. More specifically, $b_c^t (A=Requesting) \triangleq a+step*t/L$ where $t \, (0 \leq t \leq L)$ is current time, $L$ is the window length for each segment, $a$ and $step$ are offset and normalization constants, respectively (set to $a=0.4$ and $step=0.2$). The complement value then defines the BBA for the other state: $b_c^t (A=Operating)  \triangleq 1-b_c^t (A=Requesting)$. As time moves on, the probability of instrument request increases linearly over time. 

\section{Experiment} \label{sec:exp}
To validate the performance of the proposed turn-taking prediction algorithm, a number of  experiments were conducted. In the following, the experiment setup, associated evaluation schemes and metrics are first described (section \ref{sec:exp-setup}). These metrics are used to evaluate the performance of: 1) automatic feature selection (section \ref{sec:exp-feature}); 2) LSTM vs DTW for early prediction (section \ref{sec:exp-early}); 3) DST fusion methods (section \ref{sec:exp-dst}); 4) algorithms vs human's performance (section \ref{sec:exp-human}). 

\subsection{Experiment Setup and Evaluation Scheme} \label{sec:exp-setup}
This section describes the general experiment setup and the evaluation metrics for all the following experiments. The aim of all following experiments is to evaluate the performance of the turn-taking prediction algorithm. The general evaluation scheme is machine learning driven, where the data set collected from section \ref{sec:hh} is used to train and test the proposed turn-taking prediction algorithm. The metrics used to evaluate the performances of the turn-taking prediction algorithm are also described below. 

\subsubsection{Experiment Setup}
\replaced[id=r1,remark={r1c10: why leave-one-subject-out?}]{to evaluate the generalization capability of the trained system, the}{all the following machine learning experiment setup follows a} \textit{leave-one-subject-out (loso)} cross validation (cv) scheme was used \cite{esterman2010avoiding}. Under \textit{loso}, in each of the $R$ folds (here $R=12$ since we have $12$ subjects' data), the data from a single subject was left out as testing while the rest $R-1$ subjects' data used as training. Under such scheme, the training and testing split never contains data from the same subject, so that the generalization capability of the proposed algorithm can be tested. \added[id=r1,remark={r1c10: why leave-one-subject-out?}]{Under such cross validation scheme, the performance of the trained algorithm when working with novel subjects' data is evaluated. This scenario is very common when deploying trained systems into actual usage, thus \textit{loso} has been commonly used when evaluating machine learning algorithms related to humans \cite{bartlett2003real, zheng2006facial, lucey2010extended,ohn2014hand}}. There were a total of $846$ segments of \textit{requesting} state and $1305$ segments of \textit{operating} state. The \textit{operating} states were segmented to have a duration equal to the median length of all \textit{requesting} states (which is approximately $40$ frames, or $2$ seconds). \par 

\added[id=r1, remark={r1c3:how was number of participants arrived?}]{The number of participants ($R=12$) was determined based on power analysis. To reach a power level of $0.8$ (i.e., $\beta=0.2$) for differentiating DTW and LSTM under \textit{TE_10} feature set, a two-sample t-test power analysis requires 56 observations for each algorithm ( pooled variance, sample mean difference as the detected difference, alternative hypothesis is inequality and $\alpha=0.05$). 12 subjects with 5 trials for each subject leads to 60 observations for each algorithm to reach this power requirement.} \par 

The hyper-parameters for the LSTM algorithm consisted of training iterations ($100,000$), learning rate ($0.001$), batch size ($128$) and number of hidden units ($32$). The LSTM training and testing was obtained through the Tensorflow library \cite{abadi_tensorflow:_2015}.

\subsubsection{Evaluation Scheme and Metrics}
to test the algorithm's performance in early prediction instead of recognition, only the beginning fraction of the unknown action segment is used to infer its class. This can evaluate the performance of predicting the action type before the action is completely finished. The fraction percentage is characterized by the parameter $\tau$, which is referred as ``percentage of the complete action/event'' in the following content. Given an unknown segment $X_i \in \mathbb{R}^{L_i \times m}$ of an event with window length $L_i$, its class $\hat{y}_i^\tau$ (\textit{requesting} or \textit{operating}) is calculated for each fraction value $\tau \in \mathcal{T}$ where $\mathcal{T}=\{10\%,20\%, \ldots ,100\%\}$. At fraction $\tau$, the data during $[0,\tau L_i]$ was used for testing and the resultant data set is $\{X_i^\tau \in \mathbb{R}^{\tau L_i \times m}\}$. \par





For each fraction point $\tau$, the resultant \textit{F1 score} was calculated to evaluate the prediction performance,  
\added[id=r4,remark={r4c17:what is false positive and negatives?}]{based on the following equations. $N$ indicates the number of examples in the test split, $y_i \in \{0,1\}$ is ground truth and $\hat{y}_i^\tau \in \{0,1\}$ is the estimated class type at fraction value $\tau$:}
\begin{align}
True Positive (TP) &= \sum_{i=1}^N {{\hat{y}_i^\tau} y_i} \nonumber \\ 
False Positive (FP) &= \sum_{i=1}^N {{\hat{y}_i^\tau} (1-y_i)} \nonumber\\ 
True Negative (TN) &= \sum_{i=1}^N (1-{{\hat{y}_i^\tau}) (1-y_i)}  \nonumber\\ 
False Negative (FN) &= \sum_{i=1}^N (1-{{\hat{y}_i^\tau}) y_i } \label{eq:f1} \\ 
Precision (P) &= \frac{TP}{TP+FP}  \nonumber\\ 
Recall (R) &= \frac{TP}{TP+FN}  \nonumber \\ 
F1\; score (F_1) &= \frac{2PR}{P+R} \nonumber
\end{align}

We also defined the metrics to study how early and confident a prediction is made, based on metrics introduced in \cite{murphy_machine_2012}. \textit{Point of First detection (PoF)} measures the time (in percentage of total duration) that it takes to detect the class correctly for the first time. \textit{Point of Confident detection (PoC)} indicates the first time (in percentage of total duration) when all detections afterwards are correct. For both \textit{PoF} and \textit{PoC} metrics, lower values indicate better performances. More specifically, given an unknown segment $X_i$, its true label $y_i$ and the early prediction results $\hat{y}_i^\tau$, the \textit{PoF} and \textit{PoC} metrics are defined as follows:

\begin{align}
PoF &= \tau^* \, \text{such that} \, \hat{y}_i^\tau \neq y_i \, \text{for} \, \forall\tau < \tau^* \, \text{and} \, \hat{y}_i^{\tau^*}  =y_i \label{eq:pof}\\
PoC &= min \, \tau^* \, \text{such that}\, \hat{y}_i^\tau = y_i \, \text{for}\, \forall\tau \geq \tau^* \label{eq:poc}
\end{align}

\added[id=r4, remark={r4c19:more discussion of PoF}]{If the prediction $\hat{y}_i^\tau$ is never correct for all possible $\tau$ values, a value of 100\% is assigned to both \textit{PoF} and \textit{PoC}.} These two metrics are calculated for each testing example \added[id=r4, remark={r4c19:more discussion of PoF}]{from both \textit{requesting} and \textit{operating} states}, and the average over all examples are calculated to summarize the performance. \added[id=r4, remark={r4c19:more discussion of PoF}]{Assume that the number of positive and negative examples are the same and always predicting $\hat{y}_i^\tau = 1$ or $\hat{y}_i^\tau=0$, the average \textit{PoF} and \textit{PoC} value would be $(0.1+1.0)/2=0.55$. This forms an upper boundary (worst performance) for the evaluation of the algorithms. Notice that flipping a coin randomly would lead to \textit{PoF} of 0.198, but a poor performance of \textit{PoC} and \textit{F1 score}. }  

\subsection{Feature Construction and Selection} \label{sec:exp-feature}
The aim of this experiment is to evaluate the effect of feature construction and selection on prediction performances. To that end, we compare the prediction performance and confidences with different feature sets, named below:
\begin{itemize}
\item \textit{Raw}: using all $M$ channels of raw values obtained from the sensors (i.e. using $\tilde{X}_i \in \mathbb{R}^{L_i \times M}$, in this case $M=50$
\item \textit{TE_m}: using the $m$ most significant features selected from the statistics-based analysis on Temporally-Encoded (TE) features. i.e., using $X_i \in \mathbb{R}^{L_i \times m}$, and $m=10$, $30$ and $50$.
\item \textit{BTE_m}: same as \textit{TE_m} but using the binarized signal instead of the continuous signal as final features.
\end{itemize}

\(10\%\) of action segments were randomly drawn out of the entire data set and were used for feature selection process. The remaining \(90\%\) action segments were left for training and testing purposes. \added[id=r1, remark={r1c11:feature select data sampled from entire dataset?}]{An alternative approach would sample $10\%$ data from the training split of each fold for feature selection process. But the selected top features between the two approaches were found to be almost identical (with only a few features ranked differently). The reason is that the contribution of one subject's data out of 12 subjects' in $\chi^2$ analysis is small and negligible, thus the resultant features were almost identical. For simplicity and faster training purposes, feature selection was conducted only once based on the samples drawn from the entire dataset}. To evaluate the effect of feature selection on performance, the DST fusion method was not used here. The data from all the modalities was concatenated together and a single network was trained on all the data. We used LSTM as the only temporal prediction algorithm for a more controlled comparison. \par 

\begin{figure}[!t]
\centering
\includegraphics[width=3.2in]{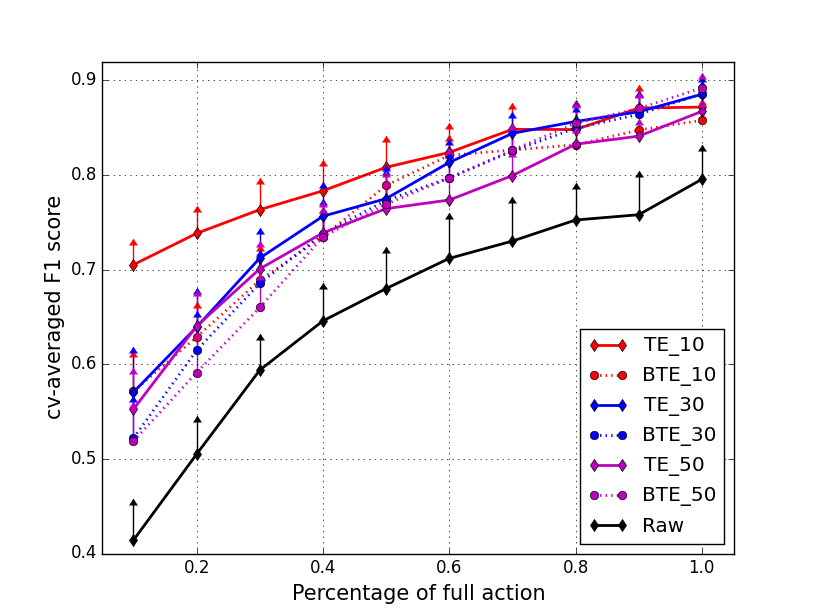}
\caption{CV-averaged \textit{F1 score} using different feature sets at each percentage point, \added[id=r1, remark={r1c12:show confidence intervals besides mean}]{arrows indicate plus one standard error.}}
\label{fig:exp-feature}
\end{figure}

The cv-averaged \textit{F1} scores is shown in Figure \ref{fig:exp-feature}, with arrows indicating plus one standard error. The different feature sets are represented by different curves in the figure. The solid lines represent the \textit{TE_m} groups while the dashed lines represent the \textit{BTE_m} groups, with the same color indicating same \textit{m} value. \replaced[id=r1, remark={r1c3:significance of F1 measure?}]{The block-design ANOVA analysis \cite{kirk1982experimental} was conducted to find out the significance of different feature selection methods. The response is the F1 score and the two factor levels are with feature selection (i.e., \textit{F1} scores with all \textit{TE_m} and \textit{BTE_m} methods) and without feature selection (i.e., \textit{F1} scores using raw features). The cross-validation fold index (i.e., $cv=\{1,...,R\}$) and fraction number (i.e., $\tau=\{0.1,...,1.0\}$) are  treated as blocks (i.e., random factors) in the model. The resultant ANOVA test shows $p<0.001$, and the Tukey's post-hoc test \cite{tukey1949comparing} shows a significant difference between the two levels (i.e. with and without feature selection, $\alpha=0.05$)} {All the feature sets using feature selection (all the \textit{TE_m} and \textit{BTE_m} curves) outperform the \textit{Raw} feature group (in black), showing the advantage of using the proposed feature selection and construction method.}

\begin{figure}[!t]
\centering
\includegraphics[width=3.2in]{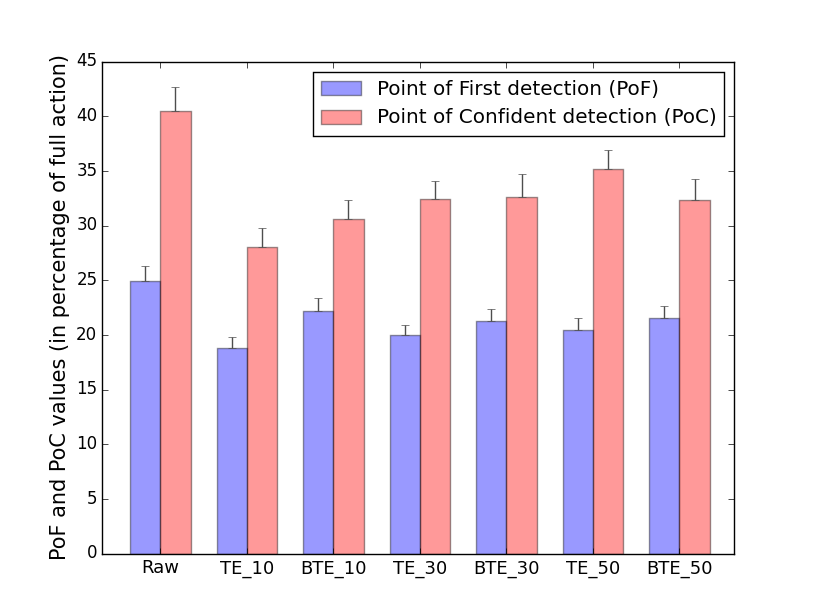}
\caption{\textit{PoF} and \textit{PoC} for each feature set. Lower values indicate better performance. Ticks on bar indicate plus one standard error.}
\label{fig:exp-feature-pof}
\end{figure}

To examine the performance of early prediction for each feature set, we present the average \textit{PoF} and \textit{PoC} for each feature set, as shown in Figure \ref{fig:exp-feature-pof}. As we can see, all the \textit{TE_m} and \textit{BTE_m} groups have smaller \textit{PoF} and \textit{PoC} values than \textit{Raw}, indicating a better early prediction performances. The \textit{TE_10} group achieves the best \textit{PoF} and \textit{PoC} scores among all the groups. \added[id=r4, remark={r4c21:TE_10 achieves best PoC?}]{As shown by Figure 8, \textit{TE_10} can achieve the highest \textit{F1 score} when little input is given (i.e., when $\tau<0.5$), leading to an accurate early and confident detection, therefore higher \textit{PoF} and \textit{PoC} results are achieved.}\par 

Notice that adding more features lead to better classification scores when looking at the complete event (i.e., corresponding to the right-most points at $\tau=100\%$ in Figure \ref{fig:exp-feature}). But it does not guarantee the optimal performance in early prediction (low $\tau$). \replaced[id=r4, remark={r4c37:over-fitting statement does not make sense}]{One potential reason is that a simpler model (i.e., a smaller value of $m$) can generalize better to partial input, thus achieving better early prediction performances.}{Such behavior of achieving optimal performance when the complete action is finished, but inferior performance when little action is observed can be explained: adding extra features in the model generates a more complex model, which is more easily to over-fit and does not generalize well to partial inputs. On the contrary, the \textit{TE_10} feature set is very accurate in early prediction, and can achieve comparable performance with the whole feature set in later stages.} TABLE \ref{tab:top-features} shows the 10 selected features in the \textit{TE_10} feature set along with their $\chi^2$ statistics. The $\chi^2$ statistics is the Pearson's cumulative test statistics, which is calculated based on expected frequency and observed frequency. The larger the $\chi^2$ statistics is (for a fixed degree of freedom), the more correlated the feature is with the action labels. For this degree of freedom ($df=1$), the upper-tail critical values of $\chi^2$ distribution to achieve $99.9\%$ significance level is \(10.828\), and all the shown statistics greatly surpass that number, indicating a high correlation of selected features with ground truth labels. 

\begin{table}[!t]
\renewcommand{\arraystretch}{1.2}
\caption{Selected Top Features}
\label{tab:top-features}
\centering
\begin{tabular}{|c|l|c|}
\hline
Rank & Feature name + Filter name & $\chi^2$ statistics\\
\hline
1 & Epoc.gyro_y + identity & 1479.2 \\
\hline
2 & Epoc.gyro_y + gabor1 & 1456.6 \\
\hline
3 & Epoc.gyro_y + gabor2 & 1430.9 \\
\hline
4 & kinect.audioConfidence + gabor1 & 1424.7 \\
\hline
5 & kinect.audioConfidence + identity & 1408.5 \\
\hline
6 & kinect.audioConfidence + gabor2 & 1388.0 \\
\hline
7 & myo.orientation_x + gabor1 & 990.3 \\
\hline
8 & myo.orientation_x + gabor2 & 975.9 \\
\hline
9 & myo.acceleration_y + gabor1 & 975.1 \\
\hline
10 & myo.acceleration_y + gabor2 & 971.1 \\
\hline
\end{tabular}
\end{table}

\subsection{Early Prediction} \label{sec:exp-early}
The goal of this experiment is to compare the performance of traditional temporal modelling algorithms (MD-DTW) with recurrent neural network algorithms (LSTM network). For comparison purposes, the DST fusion algorithm is not used. \replaced[id=r1,remark={r1c14:comparison of DTW with LSTM is unconvincing since only \textit{TE_10} is used}]{The raw feature, together with the \textit{TE_m} feature sets for $m=10,30,50$ are used to compare LSTM against MD-DTW.}{We used the feature set \textit{TE_10}, which resulted in the highest prediction performances based on the previous results.} \par 

In order to achieve early prediction, the MD-DTW and LSTM algorithm have different decision schemes. Given an unknown segment $X_i \in \mathbb{R}^{L_i\times m}$ and an evaluation fraction $\tau \, (0<\tau <1)$, MD-DTW makes a decision by comparing the distances between the beginning $\tau$ fraction of the unknown segment and the \textit{template} $\{X_*^1,X_*^0\}$, following: 

\begin{equation}
\label{eq:dtw_test_tau}
\hat{y}_i = \argmin_{y\in\{0,1\}} DTW(X_i([0,\tau L_i], X_*^y)
\end{equation}

\begin{figure}[!t]
\centering
\includegraphics[width=3.2in]{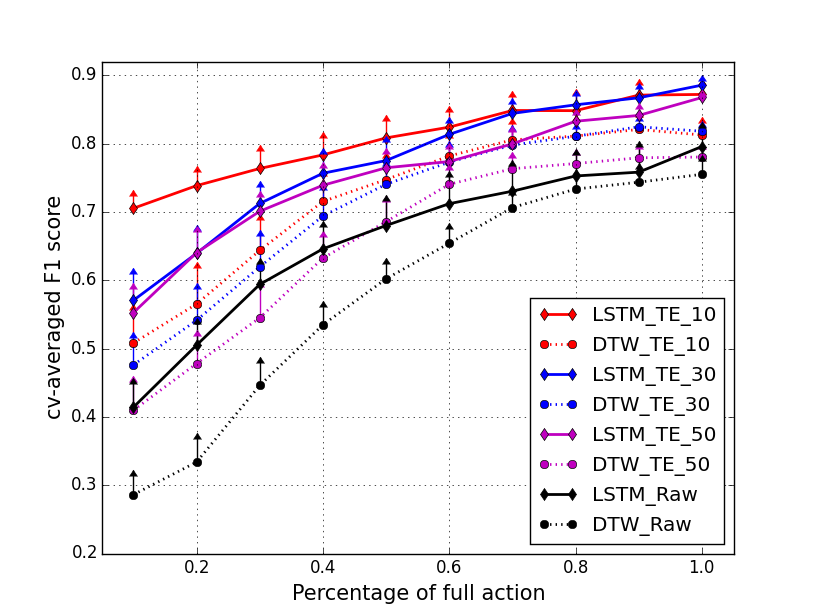}
\caption{CV-averaged F1 score using different algorithms (LSTM or DTW) with different feature sets (differentiated by color), \added[id=r1, remark={r1c12:show confidence intervals besides mean}]{arrows indicate plus one standard error.}}
\label{fig:exp-lstm-dtw}
\end{figure}

The LSTM network, alternatively, predicts the class label based on the output of memory cell at time $\tau L_i$ (instead of from the last timestamp $L_i$). The cv-averaged \textit{F1 scores} for each algorithm is shown in Figure \ref{fig:exp-lstm-dtw}. \added[id=r1,remark={r1c14:comparison of DTW with LSTM is unconvincing since only TE_10 is used}]{The solid lines represent LSTM performances and the dashed lines represent DTW performances, with the same color indicating same feature set. As shown by the image, each LSTM curve outperforms the corresponding DTW curve at each percentage point.} 
\added[id=r1, remark={r1c3:significance of F1 measure?}]{The block-design ANOVA analysis was conducted. The response is F1 score and the two factor levels are using LSTM and using DTW. The feature sets (i.e., Feature Set=\{\textit{Raw}, \textit{TE_10, TE_30, TE_50}\}), cross-validation fold index and fraction number are all treated as random factors. The resultant ANOVA test shows $p<0.001$ and the Tukey's post-hoc test shows a significant difference between LSTM and DTW ($\alpha=0.05$).} 

\added[id=r1,remark={r1c14:comparison of DTW with LSTM is unconvincing since only TE_10 is used}]{The \textit{PoF} and \textit{PoC} are also better when using LSTM compared to using the corresponding DTW algorithms. Taking TE_10 as an example, the average \textit{PoF} is $22.24$ for DTW and $18.84$ for LSTM, and the average \textit{PoC} is $32.83$ for DTW and $28.03$ for LSTM.} Therefore, it can be concluded that LSTM outperforms DTW in all metrics, including \textit{F1 score, PoF} and \textit{PoC}. These findings support the superiority of the LSTM algorithm against a more traditional temporal modelling algorithm, namely DTW.

\subsection{DST Fusion Performance} \label{sec:exp-dst}
In this experiment we evaluated the effect of adding context cues and previous decisions using the DRC fusion algorithm. The baseline condition only uses multimodal signals without any contextual cues or historic information. As comparison, the contextual cues and previous decisions are added to the multimodal signals using the DRC fusion technique, and their performances are evaluated.  \par 

Next we describe several fusion configurations which were evaluated in this experiment. The \textbf{first} configuration (\textit{TE_10}) only fuses decisions from current multimodal signals ($b_e^t, b_k^t \, \text{and} \, b_m^t$). The \textbf{second} configuration (\textit{TE_10 + context}) adds the contextual information $b_c^t$ to all the multimodal inputs. The \textbf{third} configuration (\textit{TE_10 + prev}) adds the previous decision $\tilde{b}^{t-1}$ to all the multimodal inputs. And last, the \textbf{fourth} configuration (\textit{TE_10 + prev + context}) includes multimodal signal, contextual cues and previous decisions. For all configurations in the DRC fusion process, the same weights were assigned to all the input decisions. \par 

\begin{figure}[!t]
\centering
\includegraphics[width=3.2in]{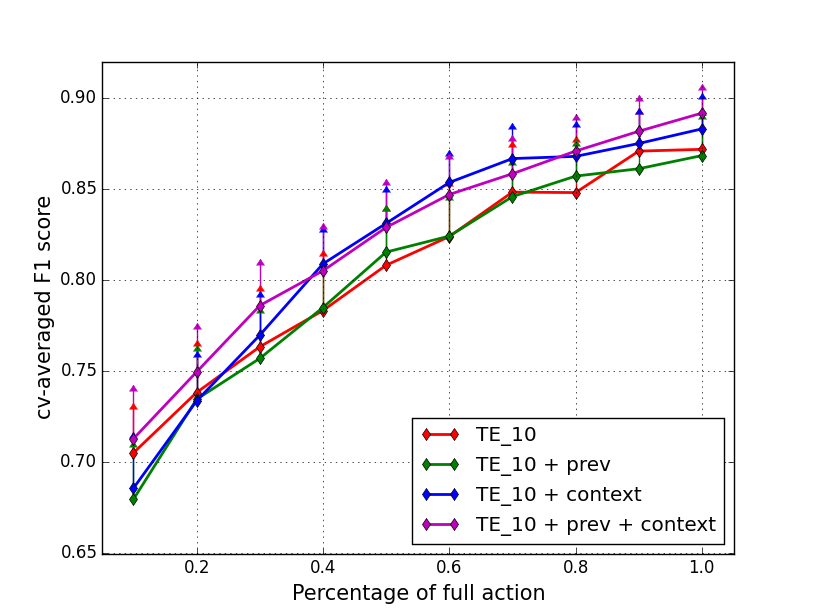}
\caption{Comparison of using multimodal signal only (\textit{TE_10}), adding context (\textit{TE_10 + context}), adding previous decisions (\textit{TE_10 + prev}), and all (\textit{TE_10 + prev + context}). Each “+” indicates a DRC fusion process. \added[id=r1, remark={r1c12:show confidence intervals besides mean}]{Arrows indicate plus one standard error.}}
\label{fig:exp-dst}
\end{figure}

Figure \ref{fig:exp-dst} shows the cv-averaged \textit{F1 score} for all 4 different configurations. 
\added[id=r1, remark={r1c3:significance of F1 measure?}]{The block-design ANOVA analysis was conducted, where response is F1 score and four factor levels are the four different DRC configurations. Cross-validation fold index and fraction number are treated as blocks same as before. The resultant ANOVA test shows $p=0.005<0.01$ and the Tukey's post-hoc test shows that \textit{TE_10 + context + prev} is significantly different from \textit{TE_10} and \textit{TE_10 + prev} group ($\alpha=0.05)$.} This indicates that the complete fusion set (\textit{TE_10 + prev + context}) achieves the best prediction performance. It is slightly outperformed by (\textit{TE_10 + context}) with a small margin in $\tau \in [0.4,0.7]$ ranges, but it is still the best combination due to its superior detection performance in the beginning fraction ($\tau=0.1$) and full fraction of the event ($\tau=1.0$). When comparing the best performed set (\textit{TE_10 + prev + context}) against using only \textit{TE_10}, there is an average \textit{F1 score} increment of $2.5\%$. 

\subsection{Human Baseline Comparison} \label{sec:exp-human}
\begin{figure}[!t]
\centering
\includegraphics[width=3.2in]{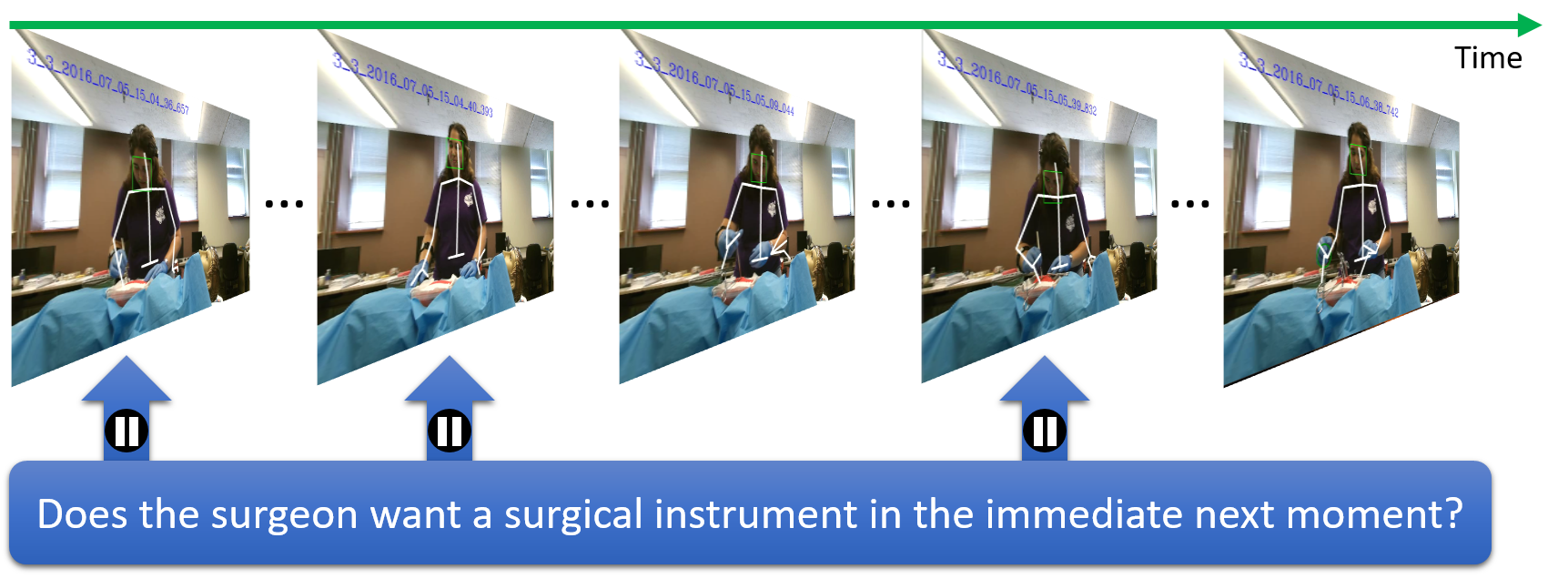}
\caption{Illustration of the human baseline acquisition process. Participants watched recorded surgery videos and answer questions about whether the surgeon wants an instrument or not.}
\label{fig:exp-video}
\end{figure}

We compared the performance of the proposed early prediction algorithm against human baseline. To that end, we recruited the same participants who took part in the data collection process (section \ref{sec:hh}). \added[id=r1, remark={r1c15:subjects are not experts}]{These participants all repeated the surgical procedure five times in order to reach a plateau in the learning curve}. The participants observed recorded surgery videos. Based on those observations they had to answer whether the surgeon wanted an instrument or not at different time periods. The illustration of the process is shown in Figure \ref{fig:exp-video}. These answers were used to compare human predictions against machine predictions. \added[id=r1, remark={r1c15:video is not appropriate}]{The video recording and replay approach is resorted to avoid interruption to the original surgery flow. It also allows evaluation of human reactions at different event fractions, since the exact duration of each turn-taking event was already spotted beforehand.} \added[id=r4, remark={r4c9:videos from real operations?}]{The recorded videos from the human turn-taking experiment (section III) were used instead of videos from real operations so that a more controlled experiment can be conducted and a fairer comparison can be made.}

Recorded videos of surgery were played to each participant under a cross-participant setting, i.e. every participant watched other's videos. The video was paused at random times, and  participants were asked to mark whether the surgeon intended to request an instrument or not. Those answers were recorded and then compared with the ground truth. The evaluation metric (\textit{F1 score}) was then calculated as human baseline performance. To determine how the video would be paused, the following procedure was followed. The video was paused within each \textit{requesting} and \textit{operating} state. Assume that a video clip contains a turn-taking event starting at $t_s$ and ending at $t_e$, this video would be paused at times $t_* \, (t_s < t_* \leq t_e)$, according to a discrete uniform distribution $t_*\thicksim$ \small $\mathcal{U}$ \normalsize $(t_s,t_e)$.  \par 

\begin{figure}[!t]
\centering
\includegraphics[width=3.2in]{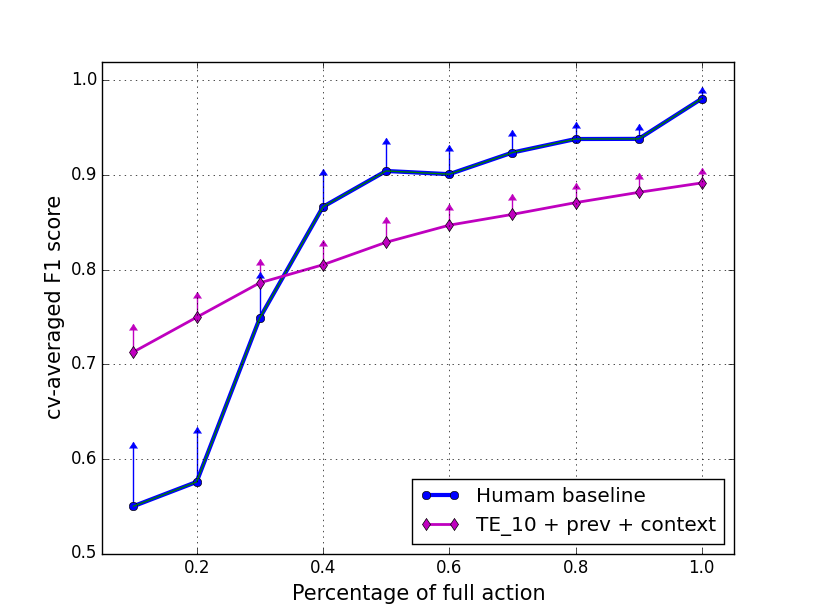}
\caption{\textit{F1 score} of LSTM compared with Human baseline. The LSTM outperforms human baseline in early stages (before \(30\%\)). \added[id=r1, remark={r1c12:show confidence intervals besides mean}]{Arrows indicate plus one standard error.}}
\label{fig:exp-human}
\end{figure}

We selected the best LSTM algorithm configuration determined from the previous experiments and compared it against the human baseline. The chosen LSTM used \textit{TE_10} feature set and the full DST fusion scheme (\textit{TE_10 + prev + context}). The performance of the human baseline, compared with the best LSTM algorithms is shown in Figure \ref{fig:exp-human}. The LSTM algorithm outperforms the human baseline for $\tau \in [0,0.3]$, and is worse than human baseline afterwards with a margin of around $6\%$. The median length of the entire action is about $2$ seconds. Therefore the proposed algorithm can deliver better prediction performance in early stages of the action (about $0.6$ seconds after the action starts, and $1.4$ seconds before the action is completed), and then \replaced[id=r4,remark={r4c41:not fair to claim comparable performance}]{is outperformed by human baseline with a margin of about $6\%$}{achieve comparable performance to the human baseline}. Such behavior grants this type of predictor the name of ``early''. 

\section{Discussion} \label{sec:duscussion}
Regarding the results of feature selection (\textbf{section \ref{sec:exp-feature}}), it has been found that higher \textit{F1 scores} are achieved when using the feature selection method compared to using the raw features. This indicates the effectiveness of the proposed feature selection process. \par 

\added[id=r4, remark={r4c42:rewrite this paragraph to be more clear}]{}
There are some observations about different \textit{encoding methods} (\textit{TE_m vs BTE_m}). Binary encoding (i.e., \textit{BTE_m}) degrades the performance compared to using multi-level encoding (i.e., \textit{TE_m}). This is due to the potential loss of information during binarization. Another observation is regarding different \textit{feature dimensions} (i.e., value of $m$). When testing complete actions (i.e., $\tau=1.0$), the performance was found to increase as feature dimension increases. However, when testing on partial inputs (i.e., small $\tau$), fewer features actually perform better. This is potentially because a simpler model (i.e., fewer interconnected nodes in the network) can generalize better to partial observations. Therefore, a trade-off exists here between having expressive models (a large $m$) to achieve high performances with complete actions, and good generalization capability to partial input (a small $m$). The best performance is achieved by balancing between these two factors, \replaced[id=r4, remark={r4c22:cannot claim optimality of $m$. In}]{and among all the $m$ values tested, an optimal value of $10$ was found. More experiments can be conducted to evaluate the optimality of $m$ values in a larger range with a more refined resolution \cite{mladenovic1997variable}.}{and we have found the optimal of $m$ to be $10$ among all the values tested in our scenario.}  

Careful analysis of TABLE \ref{tab:top-features} can shed some light on the importance of different features when predicting turn-taking. According to the table, the most significant features are mainly motion-based (encoding head, face and arm movement) and audio cues. Similar findings have been reported by \cite{saito_estimating_2015, de_kok_multimodal_2009}. \added[id=r4, remark={r4c5:surgeons turn their head?}]{Notice that surgeons in the OR have different preferences in the usage of head motions for instrument requests. Some prefer to turn their heads towards the nurse, while others prefer to focus on the operation area. Both cases are contemplated with the algorithm because of the cross-subject design.}\par


Regarding the results about comparison between DTW and LSTM algorithm (\textbf{section \ref{sec:exp-early}}), the LSTM network is found to outperform DTW in all metrics, including \textit{F1 score, PoF} and \textit{PoC}. The LSTM is one type of deep learning algorithm, and can deal with large amounts of data effectively, as has been seen in handwriting recognition \cite{graves_novel_2009}, emotion recognition \cite{wollmer_lstm-modeling_2013} and gesture recognition fields \cite{how_multiple_2014}. LSTM can model timed decision-making processes as is conducted by humans, and therefore we expect to see an analogous performance in turn-taking events. DTW was selected as the benchmark algorithm since it is still actively used in early gesture recognition \cite{izuta_early_2015} and multivariate sequence classifications \cite{mei_learning_2016}. Moreover, empirical studies have shown that DTW-based classifiers can perform at least equal (and generally better) than other classification algorithms across dozens of data sets \cite{shokoohi-yekta_generalizing_2016}. \par 

Regarding the results about DRC fusion (\textbf{section \ref{sec:exp-dst}}), it was found that adding more information channels and combining them in a rational manner (i.e., DRC) helps increasing prediction performances. Currently we have used task progress as the only context cue, including additional context information and more sophisticated sensor fusion approaches (e.g. weighted spatial-temporal DRC fusion \cite{li_visual_2013}) have the potential to further improve prediction performances. \par 

When comparing performances of the proposed algorithm vs human baseline (\textbf{section \ref{sec:exp-human}}), we have found that the proposed algorithm achieves higher \textit{F1 scores} ($12\%$ in average) than human in the beginning stage of prediction ($\tau \in [0,0.3]$). In later stages ($\tau \in [0.4,1]$), we observed a higher \textit{F1 score} ($6\%$ in average) in humans than the proposed algorithm. Thus, the algorithm's relative strength when compared with humans occurs when only partial observation is given and early prediction results are preferred. \added[id=r4, remark={r4c11:why called early prediction?}]{The proposed algorithm can clearly predict the incoming turn-taking intention much earlier than humans do, thus providing early prediction capability instead of just classification. }\par 

Putting the findings of this paper in the context of the related work yields some observations about \textit{feature importance} and turn-taking \textit{recognition performance}. Regarding \textit{feature importance} for turn-taking understanding, we have found that multimodal non-verbal signals (head, body and eye motions) are very salient in predicting turn-taking types. Similar results were found by \cite{saito_estimating_2015} (using gaze and head motion) and \cite{de_kok_multimodal_2009} (using body postures and eye movements) in the context of conversational turn-taking recognition. Regarding turn-taking \textit{recognition performance}, we have reached a \textit{F1 score} range of $[0.7, 0.9]$, depending on the prediction ranges. \cite{saito_estimating_2015} reported a similar \textit{F1 score} of $0.87$ with linear SVM for speaker's turn-taking recognition. \cite{heeman_can_2015} reported human's performance in predicting speaker's turn switches and found a \textit{F1 score} of $0.61$. \cite{gravano_turn-taking_2011} reported results in similar ranges for the same type of task, but instead using machine learning algorithms and utterance features. \cite{de_kok_multimodal_2009} studied the multi-party turn-taking prediction problem with Conditional Random Fields, but only achieved a \textit{F1 score} of \( 0.061\). \added[id=r4,remark={r4c43:task is different, so F1 is not comparable}]{The \textit{F1 scores} mentioned above are acquired under different tasks (e.g., conversational turn-taking, multi-party meetings etc) and thus cannot be directly compared to ours, but they can still show the general level of turn-taking recognition performances}. \added[id=r4,remark={r4c16:F1 score inflation?}]{In this scenario, since the number of negative instances is 1.5 times larger than that of the positive instances, the \textit{F1 score} is actually deflated by a small amount \cite{jeni2013facing}. Such data skewness issue won't bias the comparison of different feature/algorithm/fusion methods, since they all used the same unbalanced dataset and thus the \textit{F1 score} measurement is fair to all the configurations.} To summarize our findings in the context of related work, we have reached similar conclusions regarding the importance of multimodal signals for turn-taking comprehension, and achieved high performance for turn-taking prediction. In addition, we addressed the early prediction problem and presented results related to early predictability. \par 

There are some limitations of the proposed turn-taking algorithm. Currently, the proposed early turn-taking algorithm \added[id=r4, remark={r4c12:feature were pre-segmented}]{generates a decision based only on the information included in an isolated and segmented event, independently from the other nearby events}. Conversely, humans continuously gain context information about the surgical task, such as task progress, timing of previous instrument requests and emergency levels. Based on the continuous information acquisition and building, a decision is made. All such context cues contribute to the identification of turn-taking events. Future work will focus on incorporating more contextual cues in a continuous manner, mimicking more of the human behavior. 

\section{Conclusion} \label{sec:conclusion}
Early turn-taking prediction is critical for natural human-robot collaborations, since it can provide more time for robot to plan and execute its actions. Such prediction capability would induce a more fluent and smooth turn-taking transition. This paper discusses an approach consisting of an early turn-taking prediction algorithm for human robot collaboration using multimodal signals. Specifically, the design and implementation of a turn-taking prediction algorithm for a robotic scrub nurse system is presented. The proposed algorithm is evaluated on a simulated surgical procedure data set, and is found to be superior to its algorithmic counterparts, and is better than the human baseline when little partial input is given (less than \(30\%\) of full action). When given only \(30\%\) of full action (\(0.6\) seconds after the action starts and \(1.4\) seconds before the action finishes), our algorithm achieves a \textit{F1 score} of \(0.80\). After the beginning stage, our algorithm can achieve comparable performances with humans as the action progresses, reaching a \textit{F1 score} of \(0.90\) when observing the entire action. Such performances justify that the proposed early turn-taking prediction algorithm can reach high prediction performances early in the turn-taking process, granting robots more time to plan and execute their motions. Specifically, when the robot serves as an assistant to the surgeon, planning time is required to figure out where the surgical instruments should be placed for prompt delivery. When extending these concepts to other scenarios, early turn-taking prediction can also shed light on the design of other proactive and anticipatory robotic systems. \par 

Future work \added[id=r2,remark={r2c2:more comprehensive human states?}]{includes proposing a more comprehensive human state definition}, giving a richer spatial-temporal feature construction method and including more contextual information to improve the early prediction algorithm. \added[id=r1,remark={r1c7:confusion about feature construction and selection}]{The runtime dynamic feature selection capability is also going to be investigated.} \added[id=r4, remark={r4c44}]{Methods on how to integrate the turn-taking prediction results with the proactive robot motion planning will also be explored.} We also plan to validate the proposed turn-taking prediction algorithm \added[id=r4, remark={r4c1:clinical validation?}]{with clinical faculty in the operating room}.

\section*{Acknowledgment}
The authors would like to thank Dr. Rashid Mazhar and Dr. Carlos Velasquez from Hamad Medical Corporation (Qatar) for their collaboration and discussion of the project. That authors would also like to thank all the members of ISAT lab for their inspiring discussions.   

\ifCLASSOPTIONcaptionsoff
  \newpage
\fi



\bibliographystyle{IEEEtran}
\bibliography{ref}

%

%

\begin{IEEEbiography}[{\includegraphics[width=1in,height=1.25in,clip,keepaspectratio]{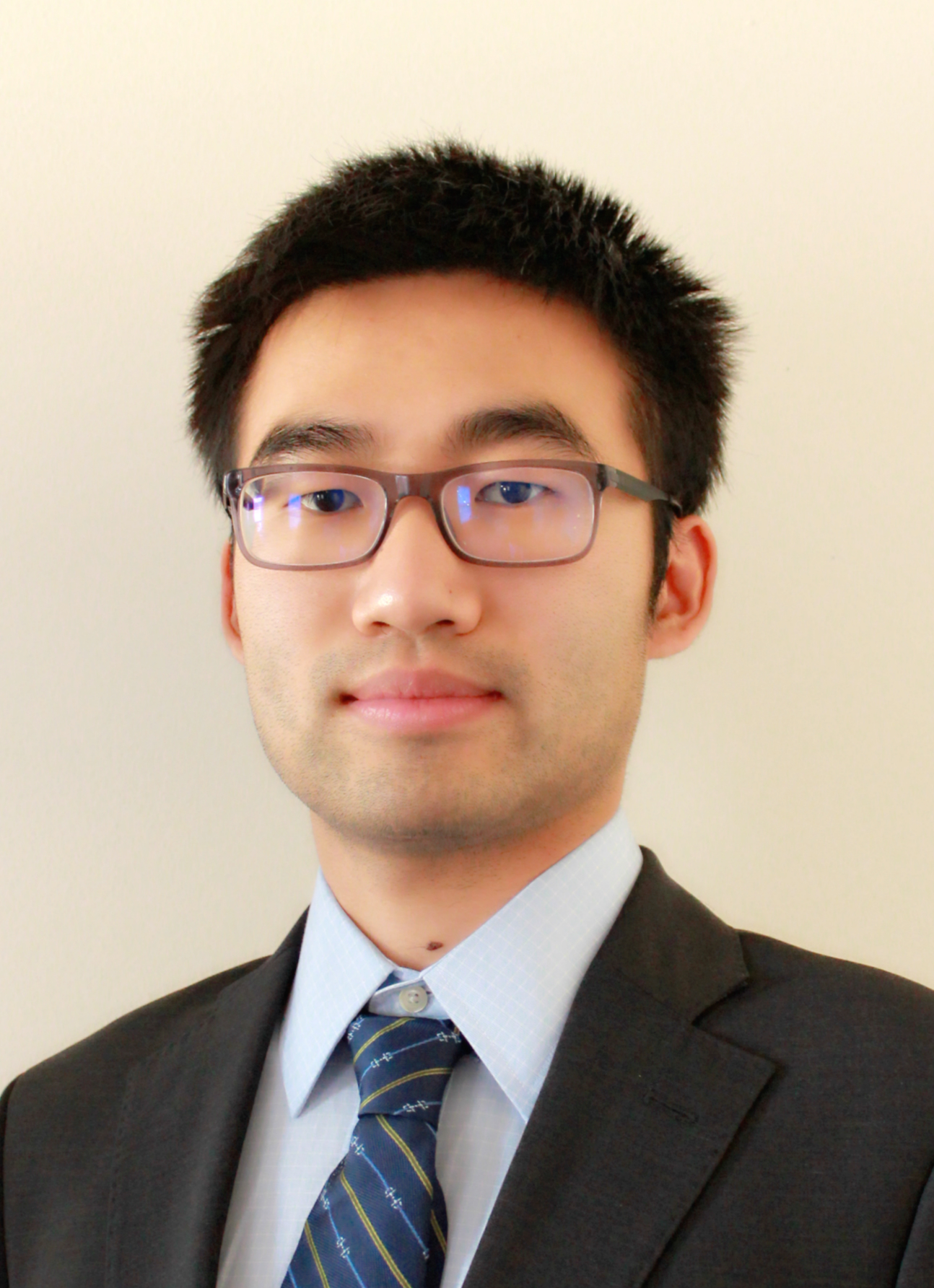}}]
{Tian Zhou}
is currently a Ph.D. student at the School of Industrial Engineering at Purdue University, working in Intelligent Systems and Assistive Technologies (ISAT) Lab. He received his M.S. in Electrical and Computer Engineering at Purdue University in 2016, and B.S. in Information Science and Engineering at Southeast University (Nanjing, China) in 2013. His research focuses on gesture recognition, machine vision, robot teleoperation and human-robot collaboration. He is the recipient of Robotics Fellowship of 2016 AAAI, and best presentation award of 2015 AAAI poster session.  
\end{IEEEbiography}


\begin{IEEEbiography}[{\includegraphics[width=1in,height=1.25in,clip,keepaspectratio]{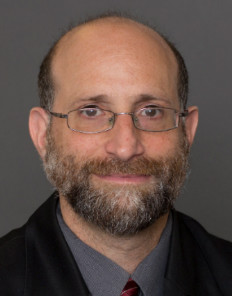}}]
{Juan P. Wachs}
is an Associate Professor in the Industrial Engineering School at Purdue University. He is the director of the Intelligent Systems and Assistive Technologies (ISAT) Lab at Purdue, and he is affiliated with the Regenstrief Center for Healthcare Engineering. He completed postdoctoral training at the Naval Postgraduate School's MOVES Institute under a National Research Council Fellowship from the National Academies of Sciences. Dr. Wachs received his B.Ed.Tech in Electrical Education in ORT Academic College, at the Hebrew University of Jerusalem campus. His M.Sc and Ph.D in Industrial Engineering and Management from the Ben-Gurion University of the Negev, Israel. He is the recipient of the 2013 Air Force Young Investigator Award, and the 2015 Helmsley Senior Scientist Fellow.  Currently he is working at UBA as part of his 2016 Fulbright U.S. Scholar Program to Argentina. He is also the associate editor of IEEE Transactions in Human-Machine Systems, Frontiers in Robotics and AI, and the Journal of Real-Time Image Processing.
\end{IEEEbiography}









\end{document}